\documentclass[preprint,12pt]{elsarticle}

\usepackage{amssymb}
\usepackage[utf8]{inputenc}
\usepackage{amsmath}

\usepackage{lscape} 
\usepackage{graphicx} 
\usepackage{afterpage} 
\usepackage{booktabs} 
\usepackage{tabularx} 
\usepackage{subcaption} 

\journal{arXiv}

\begin{document}

\begin{frontmatter}



\title{StampFormer: A Physics-Guided Material-Geometry-Coupled Multimodal Model for Rapid Prediction of Physical Fields in Sheet Metal Stamping} 

\author[label1,label2]{Jiajie Luo} 
\author[label1,label7]{Mohamed Mohamed} 
\author[label1,label3]{Osama Hassan} 
\author[label1]{Haosu Zhou} 
\author[label1]{Yingxue Zhao} 
\author[label1]{Haoran Li} 
\author[label2]{Xinrun Li} 
\author[label4]{Zhutao Shao} 
\author[label5]{Yang Long} 

\author[label1]{Nan Li\corref{cor}}
\ead{n.li09@imperial.ac.uk}

\author[label2]{Jichun Li\corref{cor}} 
\ead{jichun.li@ncl.ac.uk}

\cortext[cor]{Corresponding authors}

\affiliation[label1]{organization={Dyson School of Design Engineering, Imperial College London},
city={London},
postcode={SW7 2DB},
country={UK}}

\affiliation[label2]{organization={School of Computing, Newcastle University},
city={Newcastle upon Tyne},
postcode={NE4 5TG},
country={UK}}

\affiliation[label3]{organization={Department of Computing, Imperial College London},
city={London},
postcode={SW7 2DB},
country={UK}}

\affiliation[label4]{organization={Multi-X Solution Limited},
city={London},
postcode={E6 2JA},
country={UK}}

\affiliation[label5]{organization={Department of Computer Science, Durham University},
city={Durham},
postcode={DH1 3LE},
country={UK}}


\affiliation[label7]{organization={Department of Mechanical Engineering, Faculty of Engineering, Helwan University},
city={Helwan},
postcode={4034572},
country={Egypt}}

\begin{abstract}
Traditional sheet metal forming relies on time-consuming and expensive Finite Element Analysis (FEA) for design validation, a process that significantly prolongs design cycles. While surrogate models offer faster iteration, current approaches have limitations: scalar-based methods cannot capture comprehensive field-based FEA results, while existing image-based models often ignore the critical role of material properties by focusing solely on geometry. To address this gap, we develop a physics-guided deep learning framework, namely StampFormer, which simultaneously uses component geometry and material stress-strain responses to predict FEA outcomes. The StampFormer framework uses three core components to process data. A Material-Augmented Geometric Network (MAGN) first fuses geometric and material data. This information is then integrated at various levels by a Hierarchical Material Embedding Injection Unit (HMEIU) before being processed by the primary network backbone, an adapted Swin-UNet. We evaluated our model on the stamping of a crossmember panel with two simulation datasets for steel and aluminium panels, and results demonstrate that StampFormer provides high-fidelity predictions of critical physical fields—including thinning, major strain, minor strain, plastic strain, and displacement—in under a second. Compared with ground truth FEA, our model achieved an average relative error of less than 8.5\% on the four 2D fields and a mean squared error of less than 1.2 mm$^2$ for the 3D displacement field. In summary, we introduce a practical and efficient framework that integrates multimodal information, namely geometry and material properties, to provide fast and accurate predictions, enabling designers to perform real-time manufacturability assessments.
\end{abstract}

\begin{graphicalabstract}
\includegraphics[width=\linewidth]{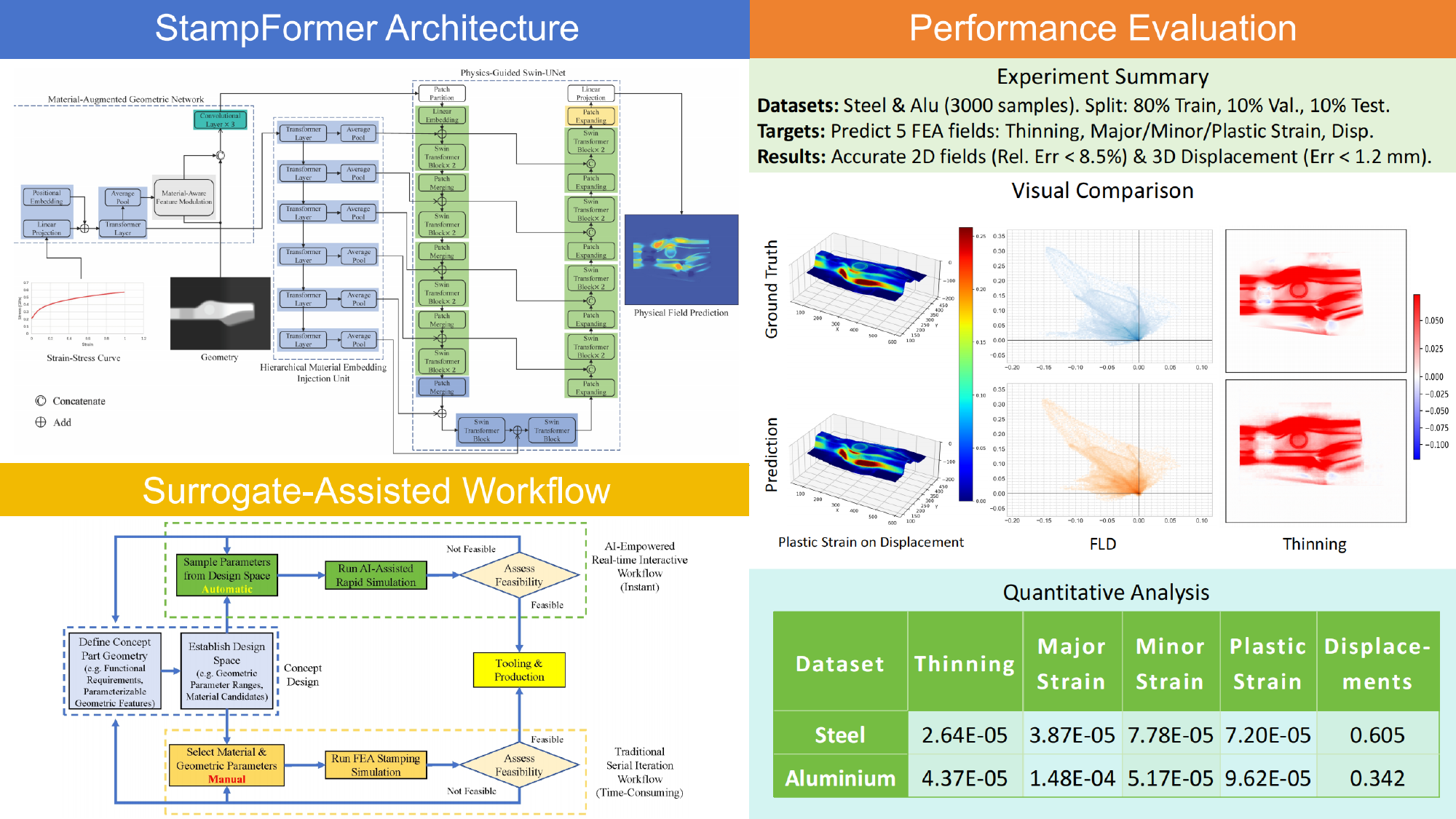}
\end{graphicalabstract}

\begin{highlights}
\item An AI surrogate model for rapid prediction of sheet metal forming fields.
\item Considers both part geometry and material properties as multi-modal inputs.
\item Provides accurate full-field results for thinning, strain, and displacement. 
\item Reduces simulation time from hours (FEA) to subsecond AI inference.
\item Enables real-time manufacturability assessment early in the design cycle.
\end{highlights}


\end{frontmatter}


\section{Introduction}
Sheet metal forming is crucial for producing complex parts in various sectors. This includes aerospace \cite{blala2023forming}, medical devices \cite{wikeckowski2022numerical}, and, notably, automotive manufacturing \cite{attar2021rapid}. As technology advances and performance demands increase, manufacturers have increasingly adopted advanced steels and aluminium alloys with high strength-to-weight ratios to produce high-performance and more fuel-efficient vehicles in recent years \cite{perka2022advanced, chandel2021review}. Despite the significant advantages these materials offer for vehicle lightweighting, it remains a challenge to design feasible and optimal parts with them \cite{hua2021investigation}. This is because their forming processes often exhibit complex behaviors, such as thinning, limited ductility, and increased springback, which pose substantial challenges to the design of manufacturable components \cite{attar2021design}. To handle this complexity, it is essential to implement a rigorous and efficient design validation process. This step is crucial for ensuring that a part is manufacturable before investing in costly tooling and production lines \cite{li2024integratedLW}.

\begin{figure}[!h]
\centering
\includegraphics[width=\columnwidth]{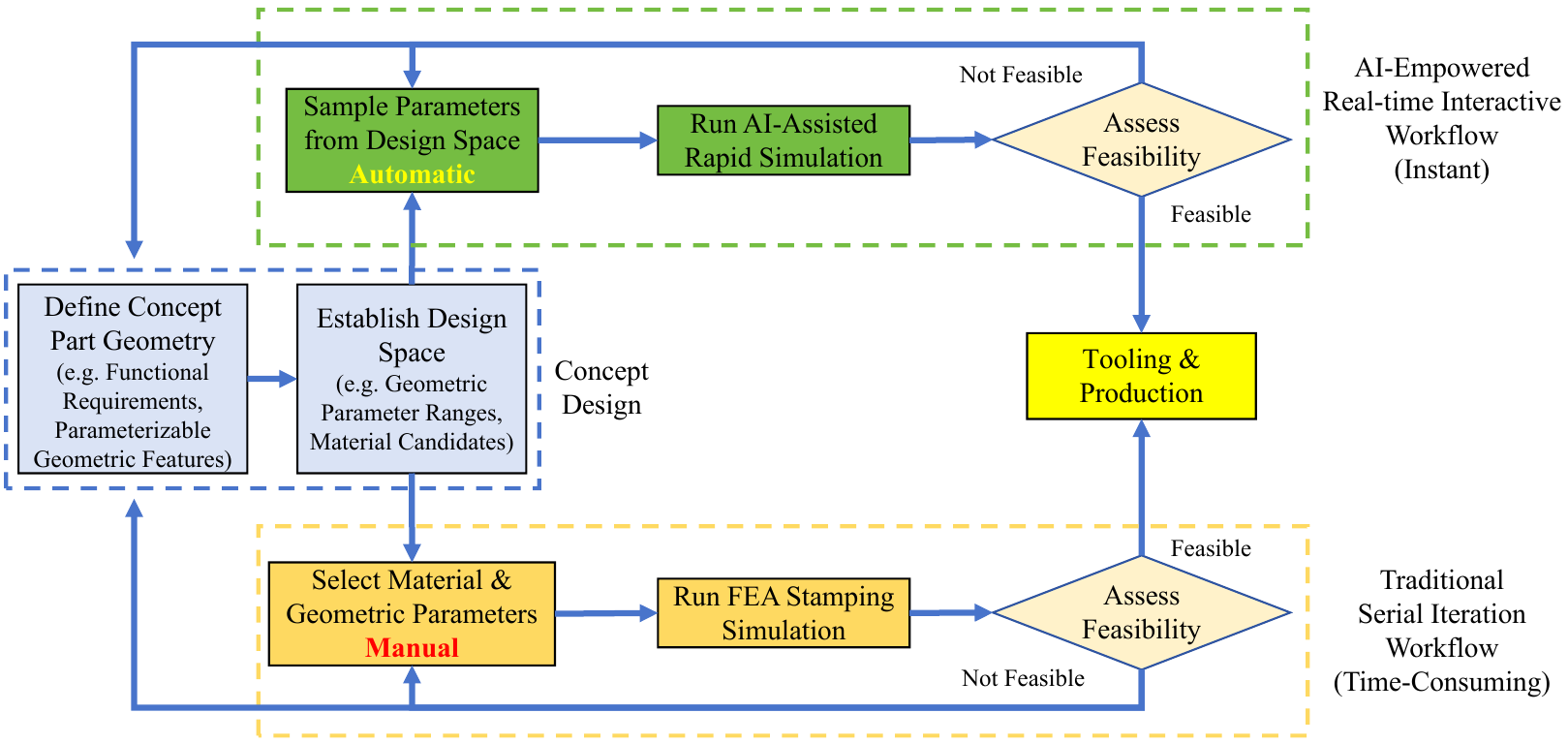}
\caption{Workflow of traditional sheet metal forming design and the proposed surrogate-assisted AI-empowered design process.}\label{fig:workflow}
\end{figure}

As illustrated in the section \textit{Traditional Serial Iteration Workflow} of Figure \ref{fig:workflow}, the conventional sheet-metal design process is constrained by a linear and repeated validation cycle \cite{alawadhi2024structural}. In this workflow, designers first create a concept part and define its geometry, and engineers then select materials and geometric parameters manually based on their experience to run a feasibility check. Such feasibility analysis relies on complex Finite Element Analysis (FEA) simulations that require significant computing resources. FEA offers precise predictions of how materials will flow, along with the potential failure regions. This makes it an essential instrument for confirming a design before final production \cite{attar2021design}. However, its steep computational cost prevents it from being an efficient tool for fast sheet metal design. In this cycle, each iteration introduces a substantial latency of hours or even days, severing the interactive link between design ideation and performance verification \cite{li2024integrated, zhou2025multi}. This inevitably lengthens the design cycle and pushes up the cost of development \cite{attar2021rapid, data_driven_innovation_2024}. 

Over the past few decades, advances in computing hardware and technology have allowed Artificial Intelligence (AI) to fundamentally change traditional engineering practices and boost productivity \cite{nti2022applications}. As a result, the application of AI in manufacturing has recently garnered significant attention in academic research \cite{tapeh2023artificial}. Some typical applications of AI in manufacturing include smart quality inspection \cite{taheri2025artificial}, predictive maintenance \cite{ucar2024artificial}, and supply chain optimization \cite{grover2025ai}. For example, Kardovskyi {\it et al.} achieved quality inspection of steel bar installation via mask R-CNN \cite{kardovskyi2021artificial}. In \cite{senoner2022using}, Senoner {\it et al.} used explainable AI to improve process quality of semiconductor manufacturing. With ongoing exploration by researchers, AI-assisted design optimization using surrogate models has surfaced \cite{plathottam2023review, attar2021rapid, zhou2022, zhou2025multi}. As shown in the AI-empowered real-time workflow in Figure \ref{fig:workflow}, AI can rapidly simulate designs to accelerate the design-validation feedback loop, replacing time-consuming FEA. However, many existing geometry-focused surrogates \cite{zhou2022, attar2021deformation} often overlook the critical influence of material properties, a gap this work aims to address.

In this paper, we introduce StampFormer, a novel physics-guided deep learning framework designed to account for the critical impact of material properties in predicting physical fields (i.e., thinning, strain, and displacement) during sheet metal forming. In any real-world forming process, a part's final performance is co-determined by its geometry and the material used. Our method introduces a multi-modal architecture that is, to the best of our knowledge, one of the first image-based sheet-forming surrogates to explicitly combine a part's geometry and its material properties (i.e., stress-strain responses) to predict FEA outcomes. This makes our model a better fit for real-world scenarios. The main contributions of this work are as follows:
\begin{enumerate}
    \item By providing rapid, low-barrier feasibility assessments, our surrogate model significantly accelerates design cycles, effectively reducing repetitive design loops. This paradigm shift empowers designers to conduct real-time manufacturability assessments of their concepts, facilitating a more agile and proactive design-to-production workflow.
    \item Our multi-modal, image-based approach surpasses the limitations of both scalar-based techniques, which fail to capture comprehensive spatial information, and conventional image-based models, which typically only consider geometric inputs while neglecting the crucial impact of material properties, providing a more holistic and physically realistic prediction of forming outcomes than existing methods.
    \item The introduced architecture achieves effective cross-modal fusion through a dual-stage integration strategy. First, the Material-Augmented Geometric Network (MAGN) enriches the initial geometric input with material-specific context; subsequently, the Hierarchical Material Embedding Injection Unit (HMEIU) injects material embeddings at multiple hierarchical levels, maintaining high-fidelity material-geometric coupling throughout the feature extraction process.
\end{enumerate}

The rest of the paper is organized as follows. Section 2 reviews related work. We detail the problem definition and how the data was prepared in Section 3. Section 4 presents the proposed network architecture, while Section 5 details the experimental setup. Section 6 presents and discusses the results. Section 7 discusses engineering implications and metric choices, and Section 8 concludes the paper.

\section{Related Work}
To address the workflow latency between the design and validation of traditional design cycles, the engineering field has increasingly turned to data-driven surrogate models. As depicted in the section \textit{AI-Empowered Real-time Interactive Workflow} of Figure \ref{fig:workflow}, these models are trained to learn the complex input-output relationships of physical simulations in near real-time, thereby enabling rapid design iterations \cite{cheng2024review, alizadeh2020managing, ling2022overview}. A relevant example can be found in structural mechanics, where surrogate models are developed to accelerate FEA. By replacing complex, time-consuming simulations with a machine learning model, researchers can rapidly predict the dynamic response, such as the displacement of a truss structure, thereby preventing the need to rerun complex simulations \cite{10143077}. Early data-driven surrogate models, like those in \cite{shi2020novel,hart2020fast,jankovivc2024data}, were relatively simple. These approaches, which we refer to as scalar-based models, worked by regressing a few key design parameters against a single performance metric to simplify complex simulation processes. 

While these scalar-based models—which predict a single, critical metric such as maximum thinning rate—offer significant speed, they suffer from a major drawback: they provide only limited output information compared to image-based surrogate models \cite{kim2020prediction,ren2020thermal,tian2021deep}. For example, in the hot stamping of a vehicle component, key variables like geometry and processing conditions influence the final thinning field. Predicting the entire thinning field with image-based machine learning methods, rather than just predicting its maximum value, provides much more comprehensive information for engineers \cite{attar2021rapid}.
Similarly, Li {\it et al.} \cite{li2024integrated} used an image-based machine-learning method to construct the surrogate model for rapid prediction of crashworthiness performance in vehicle panel components. This approach provided engineers with richer information, such as precise location, directly from the final predicted image-based outputs. While existing image-based surrogate models can provide a richer output than scalar-based surrogate models, they still have a major limitation: current architectures, while adept at handling geometric inputs, lack specialized methods for representing and processing fundamentally different data modalities. They cannot fuse 1D material property curves with 2D part geometries effectively. This limitation becomes critical in real-world applications, as material variations are known to significantly influence the final forming responses.

Solving the problem of integrating material and geometric data requires a suitable architectural foundation. The encoder-decoder structure, for example, is widely used for semantic segmentation, where the goal is to produce a dense, pixel-wise probability map, analogous to the physical fields in FEA simulations. Among the existing image-based FEA surrogate models, many have also been designed with an encoder-decoder structure \cite{attar2021rapid, li2024integrated, zhou2025multi, zhou2022}. As a pioneering encoder-decoder model, U-Net is widely employed in the field of medical image segmentation \cite{azad2024medical}. Over the past decade, encoder-decoder models for semantic segmentation have advanced considerably, undergoing multiple rounds of iteration \cite{luo2024semantic} driven by the development of architectures such as Vision Transformer \cite{kim2025systematic}, Vision Mamba \cite{liu2024vision}, and Swin Transformer \cite{kumar2025improved}. A number of superior architectures, including TransUNet \cite{zhu2024sparse}, UMamba \cite{jain2025differential}, and SwinUNet \cite{cao2022swin}, have since been introduced. Despite these architectural advancements, their application in manufacturing has yet to address the critical gap identified earlier: the need to integrate material properties with geometry. Fortunately, the recent emergence of multi-modal learning—a subfield of AI focused on processing and fusing information from fundamentally different data types (such as 1D curves and 2D images)—provides the necessary techniques to close this gap.

\section{Data Generation and Pre-processing}
\label{dataprep}
In this study, our objective was to develop an AI-assisted surrogate model capable of rapidly and accurately predicting physical fields in sheet metal forming processes. To accomplish this, it is essential to construct a comprehensive dataset that effectively captures the complex interdependencies among material properties, part geometry, and the resulting physical fields obtained from FEA simulations. The dataset for training the StampFormer model consists of three core components: material data, geometric data, and the corresponding FEA simulation results. Specifically, we need to generate input data that includes a variety of geometric designs and a wide range of material stress-strain curves.

For each combination of a geometric design and a material stress–strain curve, a corresponding FEA result is generated. This section details the generation of these raw data components and their pre-processing into the 2D and 1D formats required by the StampFormer network.

\subsection{Material Data Generation}
The predictive accuracy of the StampFormer model is highly dependent on both the quantity and quality of data used during the training, validation, and test phases. Consequently, the generation of virtual material datasets, as opposed to relying solely on experimentally measured data, is of critical importance. Virtual materials enable systematic exploration of a broad spectrum of stress–strain responses, extending beyond the limitations of experimentally tested materials. This approach ensures enhanced diversity and adequate representation of the material design space, thereby improving the robustness and generalizability of the surrogate model. For computational consistency, all generated stress-strain curves are uniformly sampled to a fixed length of $T=100$ data points.

\begin{figure*}[!htbp]
    \centering
    \subfloat[Initial dataset (101 curves).]{\includegraphics[width=0.32\linewidth]{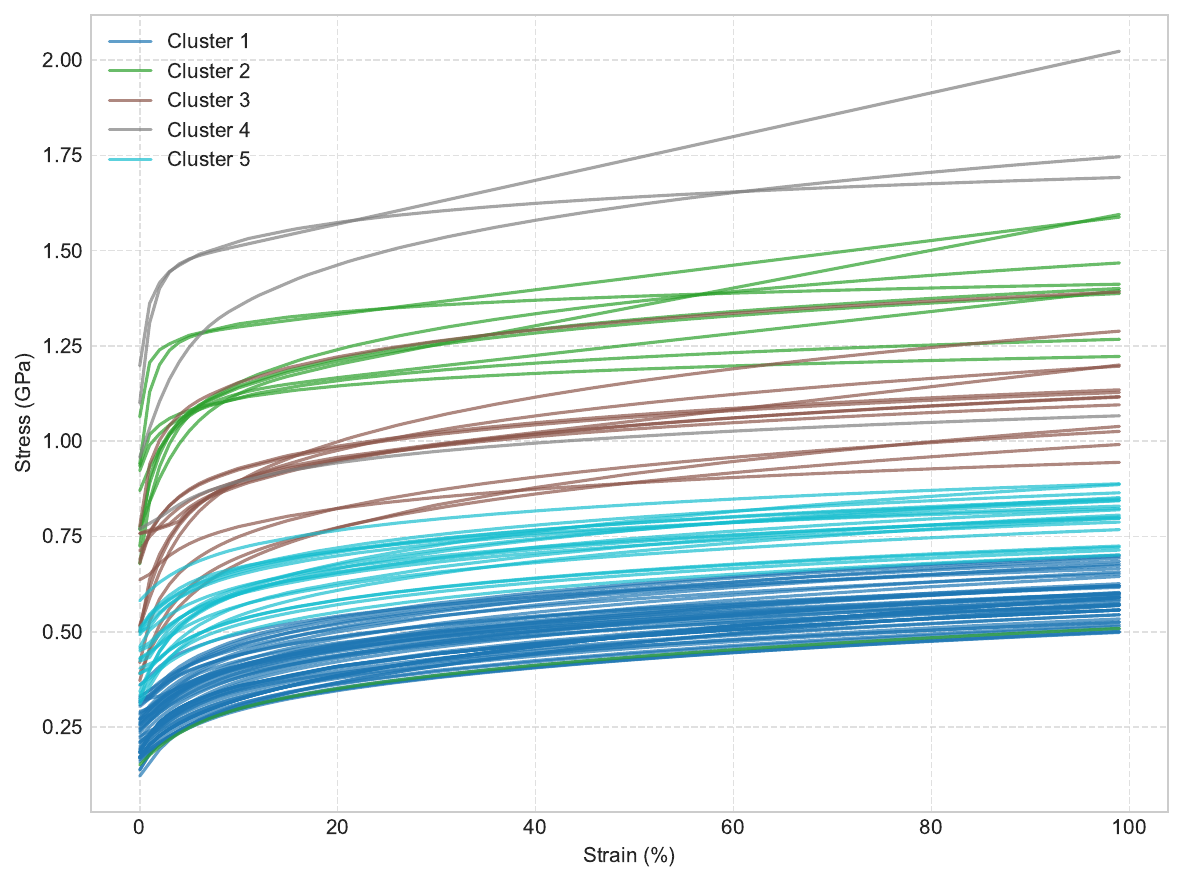}\label{fig:steel_101}}
    \hfill
    \subfloat[Intermediate dataset (200 curves).]{\includegraphics[width=0.32\linewidth]{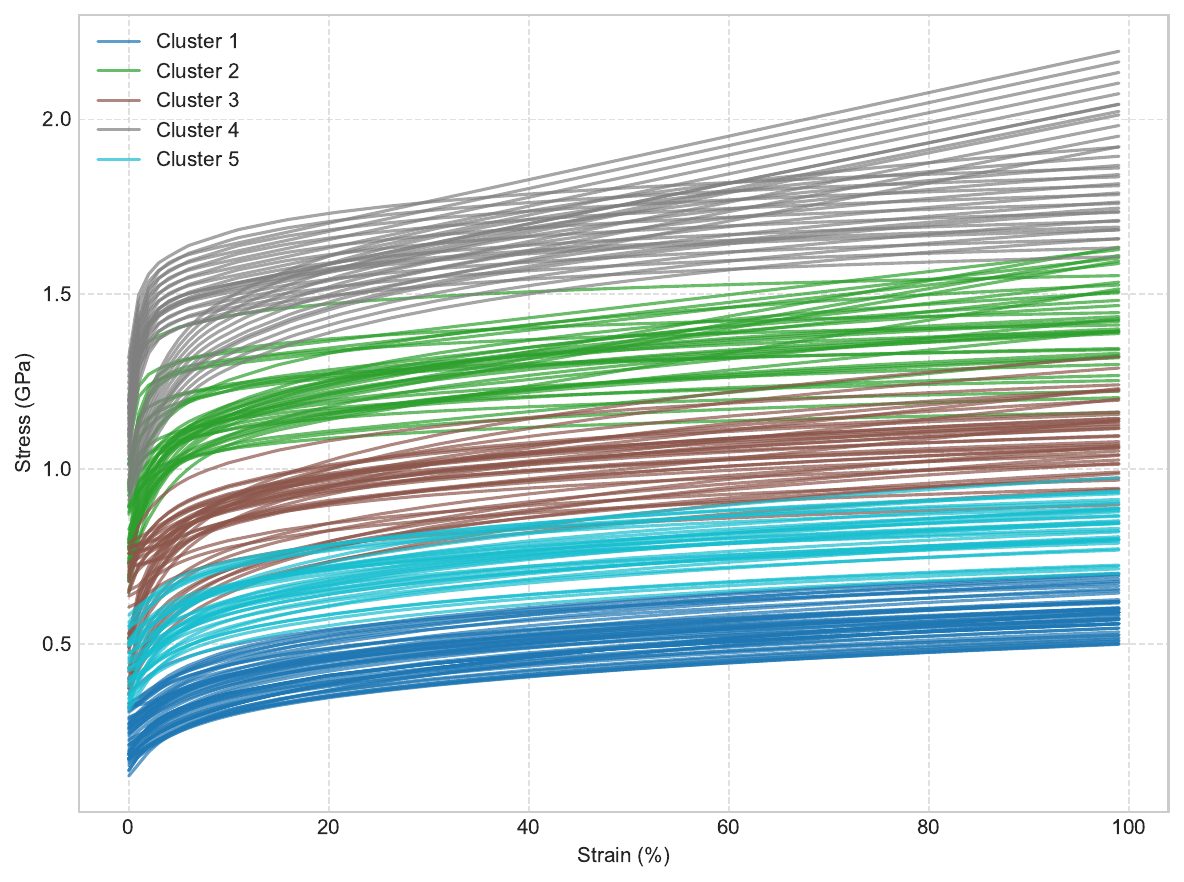}\label{fig:steel_200}}
    \hfill
    \subfloat[Final augmented dataset (600 curves).]{\includegraphics[width=0.32\linewidth]{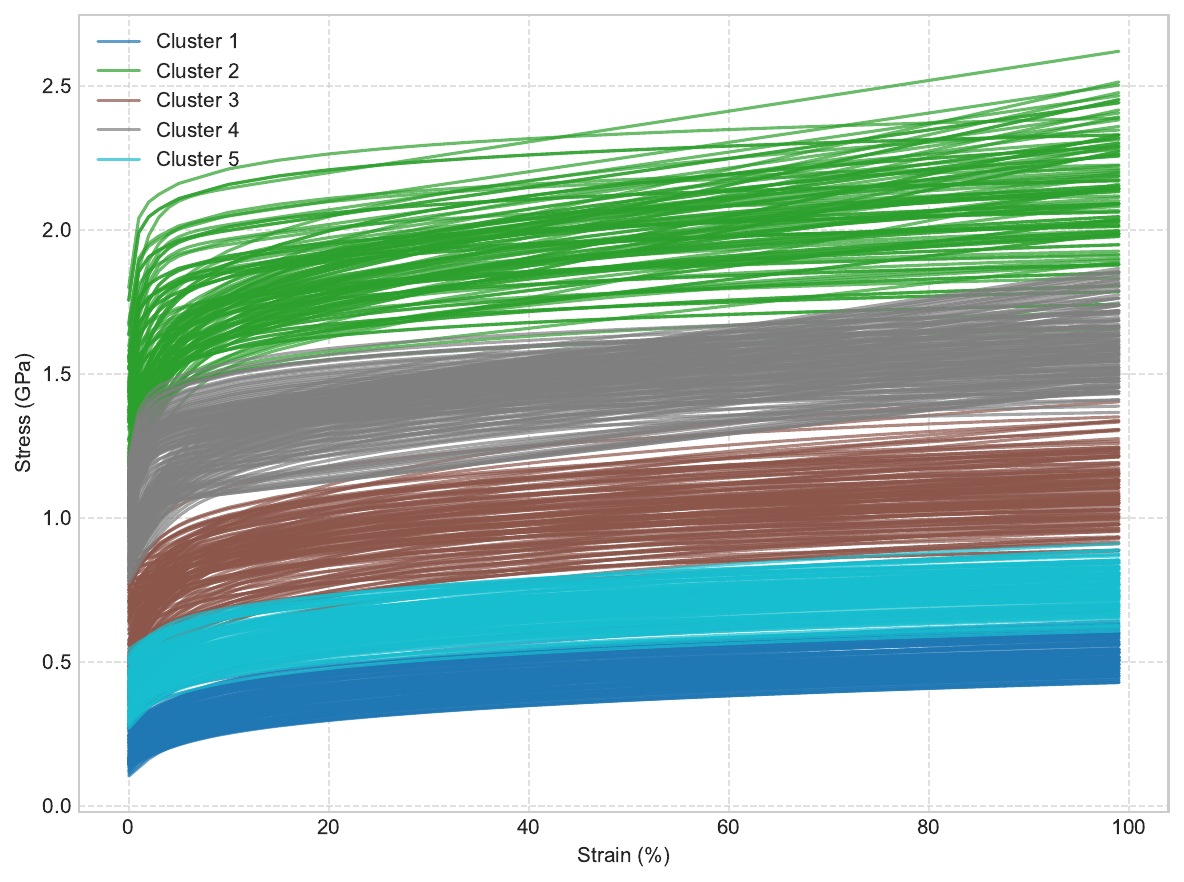}\label{fig:steel_600}}
    \caption{Illustration of the steel stress-strain dataset evolution. The dataset is expanded from its (a) initial unbalanced state of 101 curves, to (b) a uniformly balanced intermediate set of 200 curves, and finally upsampled to (c) 600 curves. All datasets are partitioned into five distinct clusters to ensure high diversity.}
    \label{fig:steel_cluster_evolution}
\end{figure*}

\begin{figure*}[!htbp]
    \centering
    \subfloat[Initial dataset (11 curves).]{\includegraphics[width=0.48\linewidth]{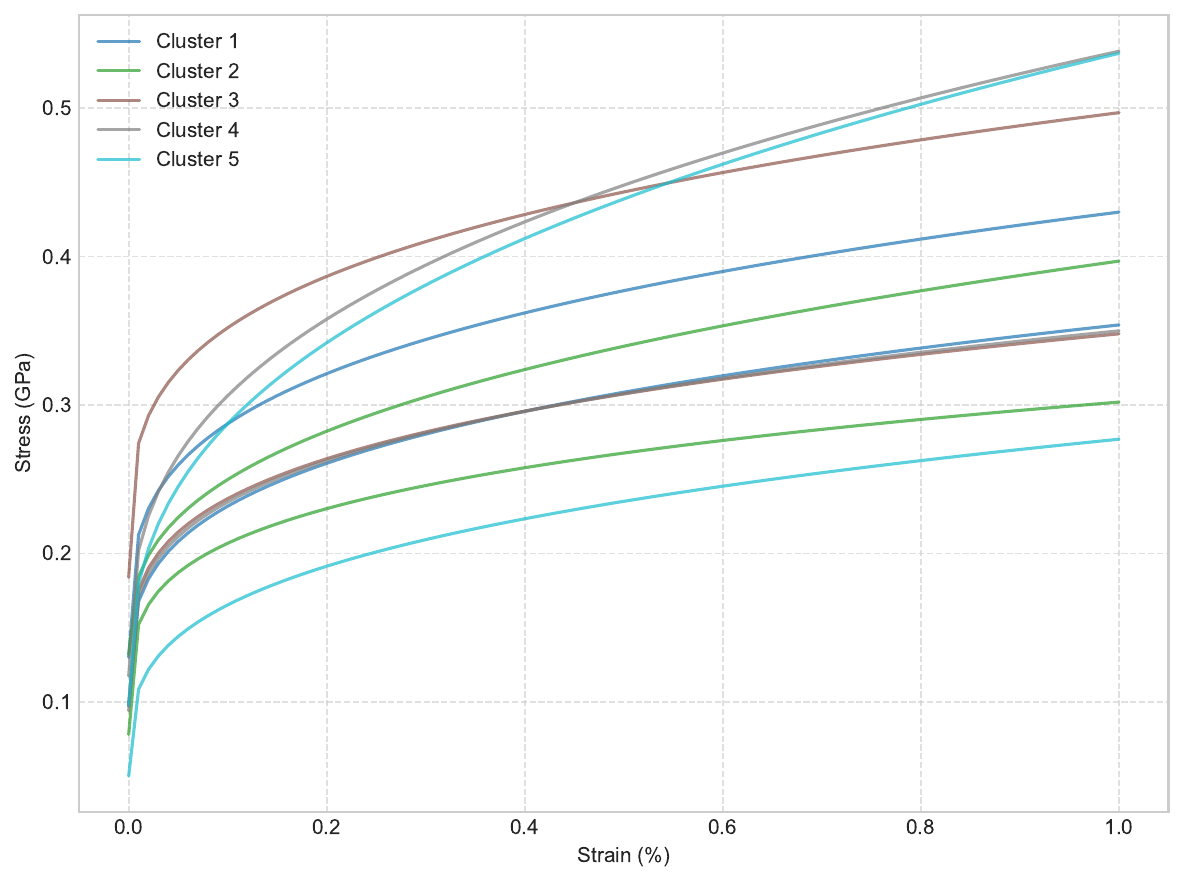}\label{fig:alu_original}}
    \hfill
    \subfloat[Final Augmented dataset (110 curves).]{\includegraphics[width=0.48\linewidth]{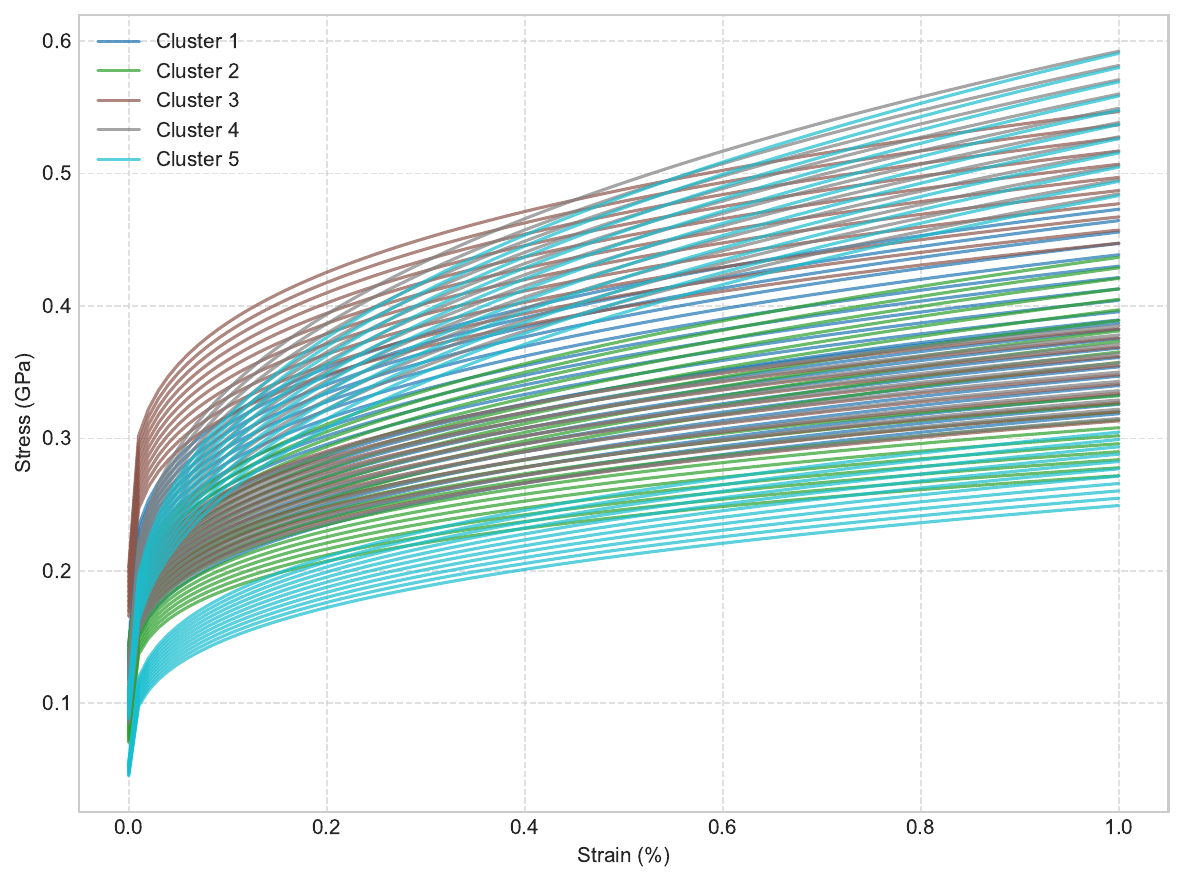}\label{fig:alu_augmented}}
    \caption{Illustration of the aluminium stress-strain dataset before and after augmentation. The (a) original 11 real-world material curves are systematically scaled to produce the (b) final augmented dataset of 110 curves, partitioned into five balanced clusters.}
    \label{fig:alu_cluster_evolution}
\end{figure*}

The initial steel dataset comprised 101 stress--strain curves sourced from an existing material database, categorized into five clusters as illustrated in Figure \ref{fig:steel_cluster_evolution}(a). The initial distribution was highly imbalanced, with cluster sizes ranging from 3 to 55 curves (e.g., Cluster 1 had 55 curves, while Cluster 4 contained only 3). To address this, a two-stage augmentation process was implemented: first, each cluster was manually expanded to 200 curves to ensure uniformity (Figure \ref{fig:steel_cluster_evolution}(b)); subsequently, systematic upsampling was applied to generate a final secondary dataset of 600 curves (Figure \ref{fig:steel_cluster_evolution}(c)). For the aluminium dataset, 11 real-world material curves (Figure \ref{fig:alu_cluster_evolution}(a)) served as the baseline. This dataset was expanded to 110 curves (Figure \ref{fig:alu_cluster_evolution}(b)) by applying multiplicative scaling factors from $\pm 10\%$ at $2\%$ increments. This procedure yielded five distinct, balanced clusters, ensuring a robust representation of material properties for model training.

\subsection{Geometric Parameter Generation}
\begin{figure}[h!]
    \centering
    \includegraphics[width=\columnwidth]{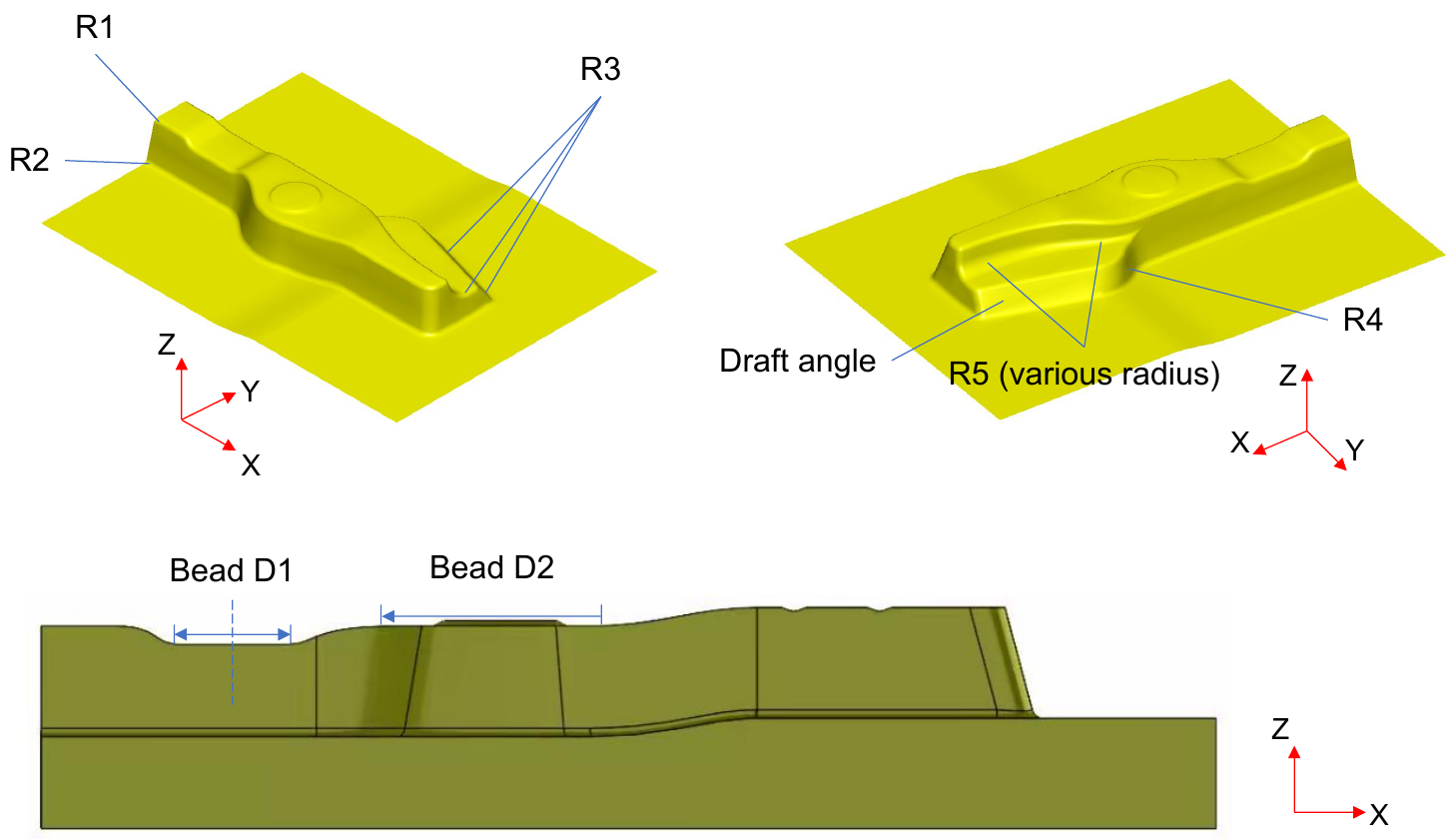}
    \caption{The key geometric features for the proposed panel.}
    \label{fig:geometry}
\end{figure}

To create a comprehensive dataset, we systematically sampled the geometric design space. As illustrated in Figure \ref{fig:geometry}, the shape of the panel includes several key geometric features, including the die radius $R_1$, punch radius $R_2$, and a draft angle. Other fillet radii, such as $R_3$, $R_4$, and $R_5$, are also introduced. The $R_5$ parameter, in particular, represents a variable fillet radius, with its starting and ending radii sampled from different ranges. These features were treated as variable parameters in this study. We used Latin Hypercube Sampling (LHS) to generate a total of 600 distinct geometric parameter sets, which define 600 unique 3D CAD geometries, with uniform coverage of the design space. The specific sampling ranges for each parameter are detailed in Table \ref{tab:geometry_params}. This method ensures efficient coverage of the design space, capturing how geometric variations—from simple to complex—affect the final forming characteristics.

\begin{table}[h!]
    \centering
    \caption{Latin Hypercube Sampling ranges for geometric parameters.}
    \label{tab:geometry_params}
    \begin{tabular}{|c|c|}
        \hline
        \textbf{Parameter} & \textbf{Sampling Range} \\
        \hline
        $R_1$ (mm) & 5 -- 10 \\
        \hline
        $R_2$ (mm) & 5 -- 10 \\
        \hline
        $R_3$ (mm) & 5 -- 10 \\
        \hline
        $R_4$ (mm) & 30 -- 60 \\
        \hline
        $R_5$ start (mm) & 5 -- 15 \\
        \hline
        $R_5$ end (mm) & 10 -- 25 \\
        \hline
        Bead D1 (mm) & 30 -- 60 \\
        \hline
        Bead D2 (mm) & 100 -- 130 \\
        \hline
        Draft angle ($^\circ$) & 35 -- 60 \\
        \hline
    \end{tabular}
\end{table}

\begin{figure}[h!]
    \centering
    \begin{subfigure}[b]{0.48\textwidth}
        \centering
        \includegraphics[width=\textwidth]{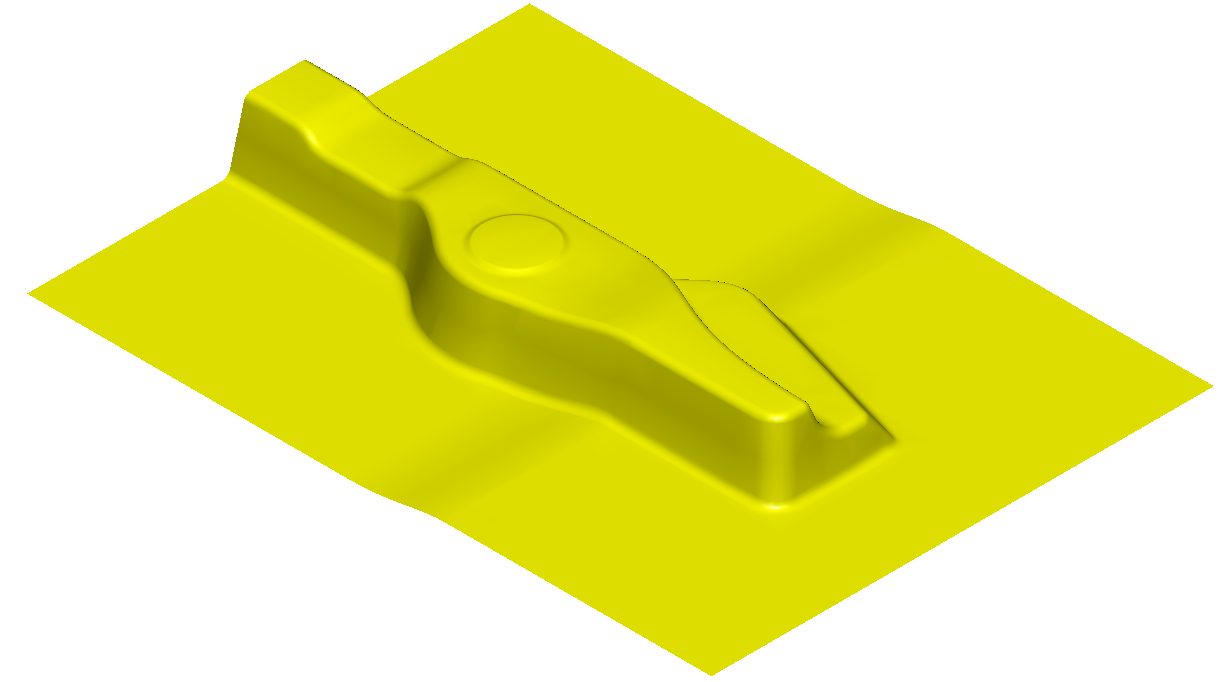}
        \caption{3D Die Geometry (Input)}
        \label{fig:geo_raw}
    \end{subfigure}
    \hfill 
    \begin{subfigure}[b]{0.48\textwidth}
        \centering
        \includegraphics[width=\textwidth]{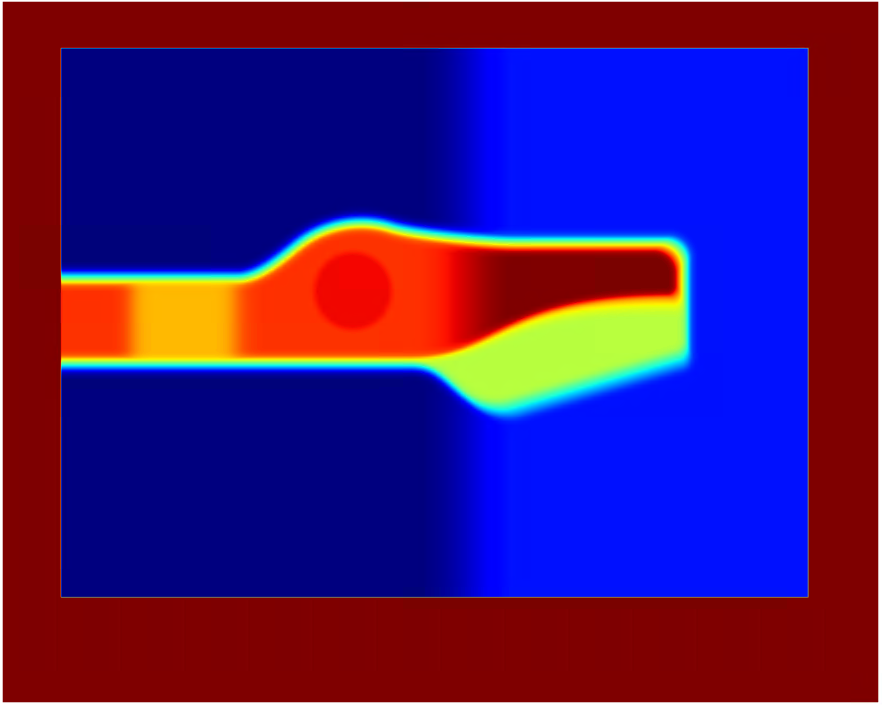}
        \caption{2D Projected Height-map}
        \label{fig:geo_proj}
    \end{subfigure}
    
    \vspace{0.5cm} 

    \begin{subfigure}[b]{0.48\textwidth}
        \centering
        \includegraphics[width=\textwidth]{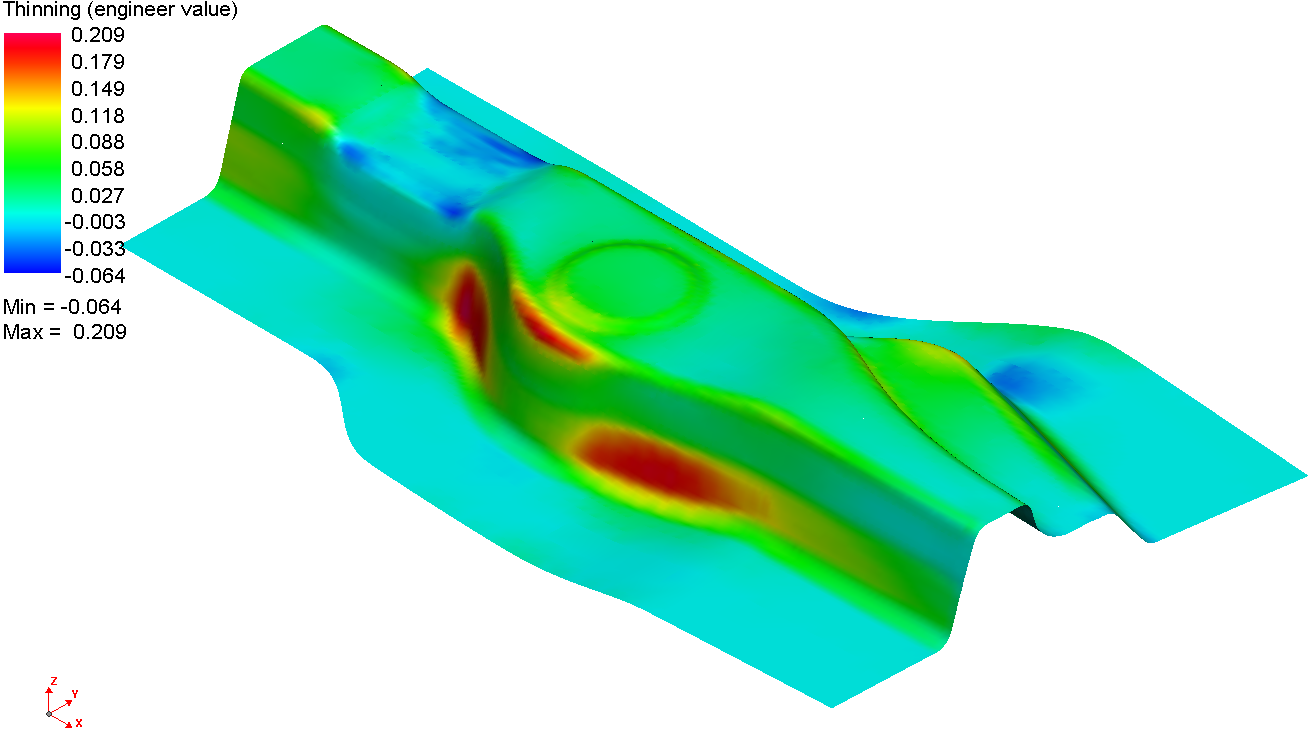}
        \caption{3D Deformed Thinning Field (Target)}
        \label{fig:field_raw}
    \end{subfigure}
    \hfill 
    \begin{subfigure}[b]{0.48\textwidth}
        \centering
        \includegraphics[width=\textwidth]{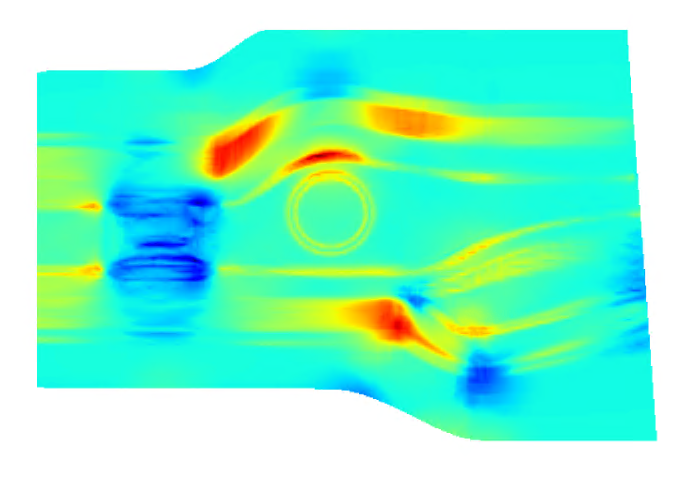}
        \caption{2D Mapped Thinning Field}
        \label{fig:field_proj}
    \end{subfigure}
    
    \caption{Illustration of the 3D-to-2D conversion process for model inputs and targets.}
    \label{fig:projection_placeholder}
\end{figure}

\subsection{Input and Target Image Preparation}
The StampFormer model accepts multi-modal inputs: a 1D vector for the stress-strain curve and a 2D image representing the part's geometry. The model outputs a 2D image representing the predicted physical field. This output is a single-channel image for 2D fields (e.g., thinning, plastic strain) or a 3-channel image for the displacement field. This necessitates the conversion of both the 3D CAD geometries and the 3D FEA simulation results into a uniform 2D image format.

The input geometry images are generated from the 3D CAD models. An information-preserving Z-plane projection is employed for the undercut-free panels considered here, as shown in Figure \ref{fig:projection_placeholder}(a) and (b). This projection relies on the assumption that the component's geometry can be accurately represented as a single-valued height-map, meaning there are no overlapping surfaces in the pressing direction. The out-of-plane height (Z-value) at each (X, Y) coordinate is recorded to create this 2D height-map. This height-map is then interpolated onto a uniform Cartesian grid with a resolution of $608 \times 768$ pixels, resulting in a single-channel image that serves as the 2D geometric input.

The target physical field images are generated from the FEA results. FEA simulations yield physical field data (e.g., thinning) on the final deformed 3D mesh as shown in Figure \ref{fig:projection_placeholder}(c). To ensure spatial alignment with the 2D input, these 3D results are mapped back to the initial undeformed 2D blank. Using the simulation's displacement field, each node on the deformed mesh is mapped to its original (X, Y) coordinate, and its corresponding physical field value is transferred. This process creates a 2D point cloud, which is then interpolated onto the same uniform grid, as illustrated in Figure \ref{fig:projection_placeholder}(d). For 2D physical fields such as thinning, major strain, minor strain, and plastic strain, this process yields a single-channel target image. For the 3D displacement field, the X, Y, and Z displacement components are similarly mapped back to the original grid, creating a 3-channel image that represents the 3D vector field.

\begin{table}[h!]
\centering
\caption{A sample of the Design of Experiments matrix for the steel dataset. Note that for each distinct geometry, one material is systematically sampled from each of the five clusters.}
\label{tab:doe_table}
\begin{tabular}{|c|c|c|c|}
\hline
\textbf{ID} & \textbf{Geometry ID} & \textbf{Material ID} & \textbf{Material Cluster} \\
\hline
0 & 472 & 104 & 1 \\
\hline
1 & 472 & 225 & 2 \\
\hline
2 & 472 & 328 & 3 \\
\hline
3 & 472 & 449 & 4 \\
\hline
4 & 472 & 532 & 5 \\
\hline
5 & 471 & 52 & 1 \\
\hline
6 & 471 & 232 & 2 \\
\hline
7 & 471 & 348 & 3 \\
\hline
8 & 471 & 412 & 4 \\
\hline
9 & 471 & 501 & 5 \\
\hline
\end{tabular}
\end{table}

\subsection{Design of Experiments}
Building upon our prepared geometry and material data, we designed experiments to generate the target FEA simulation outcomes. This process created a comprehensive Design of Experiments (DoE) matrix. For the steel dataset, we randomly partitioned the 600 geometries into training, validation, and testing sets at a ratio of 0.8, 0.1, and 0.1. Each geometry was then paired with five materials drawn from five different clusters, resulting in training, validation, and testing sets with 2,400, 300, and 300 samples, respectively. A sample of this pairing is shown in Table \ref{tab:doe_table}. The same DoE methodology was applied to the aluminium dataset. Each material system contains 3000 simulations, giving 6000 simulations in total.

\subsection{FEA Simulation Setup}
\begin{figure}[h!]
    \centering
    \includegraphics[width=\columnwidth]{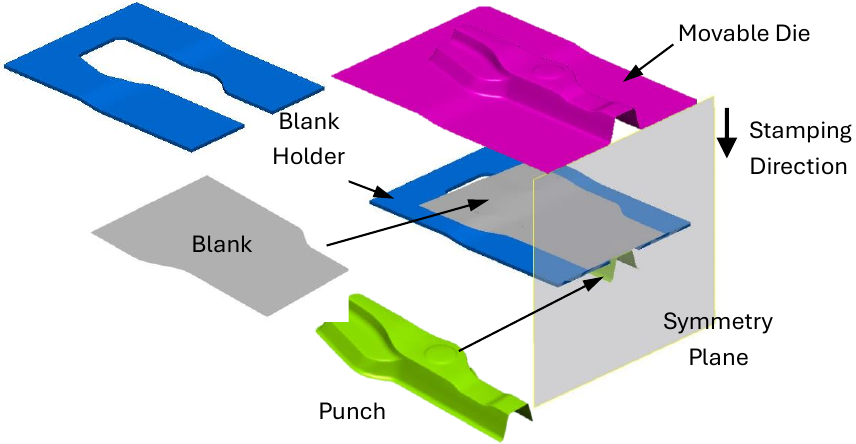}
    \caption{The FEA model setup.}
    \label{fig:FEmodel}
\end{figure}

The FEA simulations were employed to generate the training, validation, and test datasets for StampFormer. A representative automotive panel forming model was developed in PAM-STAMP (Figure \ref{fig:FEmodel}). The model integrates the tooling and blank geometries, the process setup, and the FEA solver controls. Tooling kinematics follow production practice: the die travels in the press direction toward a fixed punch, while a blank holder (binder) applies a prescribed holding pressure to regulate sheet draw-in between the binder and die. All other parameters and settings, except for material type and geometry, were fixed, as specified in Table \ref{tab:fe_setup}. The simulation matrix was generated automatically using integrated Python and PAM-STAMP scripts.

\begin{table}[h!]
\centering
\caption{The process parameters for the FEA model.}
\label{tab:fe_setup}
\begin{tabular}{|l|l|}
\hline
\textbf{Model Parameter} & \textbf{Value / Definition} \\
\hline
Blank Thickness & 1.6 mm \\
\hline
Material Model & Isotropic Lookup Table \\
\hline
Tool Type & Feasibility Stamping (3-piece tools) \\
\hline
Forming Speed & 250 mm/s (constant) \\
\hline
Blank Holding Force & 800 kN \\
\hline
Friction Coefficient & 0.12 \\
\hline
\end{tabular}
\end{table}

\begin{figure}[!htbp]
    \centering
    \includegraphics[width=\columnwidth]{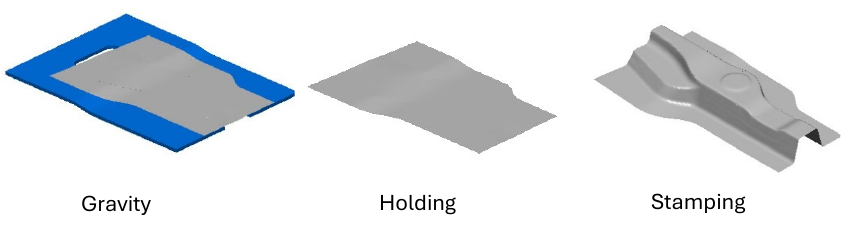}
    \caption{FE simulation stages used in the forming workflow: gravity loading, blank holding, and stamping.}
    \label{fig:simulation_flow}
\end{figure}

\begin{figure}[!htbp]
    \centering
    \begin{subfigure}[b]{0.3\columnwidth} 
        \centering
        \includegraphics[width=\textwidth]{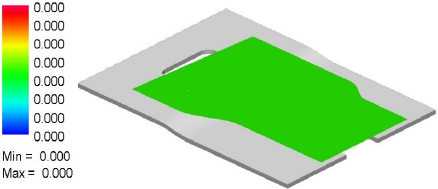}
        \caption{Gravity}
        \label{fig:sub_gravity}
    \end{subfigure}
    \begin{subfigure}[b]{0.3\columnwidth}
        \centering
        \includegraphics[width=\textwidth]{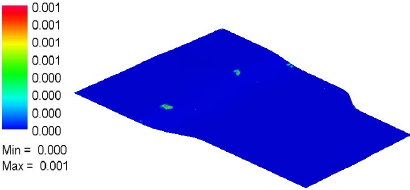}
        \caption{Holding}
        \label{fig:sub_holding}
    \end{subfigure}
    \begin{subfigure}[b]{0.3\columnwidth}
        \centering
        \includegraphics[width=\textwidth]{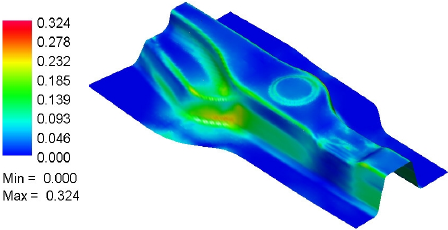}
        \caption{Stamping}
        \label{fig:sub_stamping}
    \end{subfigure}
    
    \caption{Simulation results showing the effective plastic strain distribution for one of the matrix cases.}
    \label{fig:plastic_strain_distribution}
\end{figure}

Figure \ref{fig:simulation_flow} illustrates the overall simulation flow of the stamping process for the part, which is divided into three main stages: gravity loading, blank holding, and stamping, each representing a critical step in capturing the realistic behavior of the material during forming. One example of the obtained results is the plastic strain distribution throughout the part, evaluated at the end of each stage, as illustrated in Figure \ref{fig:plastic_strain_distribution}.

\section{Network Architecture}
In this study, we introduce StampFormer, a novel deep learning framework that provides fast and accurate predictions of complex physical fields in FEA. Throughout this section, let $I_{\rm geo} \in \mathbb{R}^{H \times W \times 1}$ denote the input 2D geometry image, and let $S \in \mathbb{R}^{T \times 1}$ denote stress values sampled at $T$ fixed strain levels representing the 1D stress-strain curve. As shown in Figure \ref{fig:main_architecture}, the proposed StampFormer is composed of three main components: MAGN, HMEIU, and the Physics-Guided Swin-UNet (PGSU).

\afterpage{
    \clearpage 
    \begin{landscape}
        \begin{figure}[p] 
            \centering
            \includegraphics[width=\columnwidth]{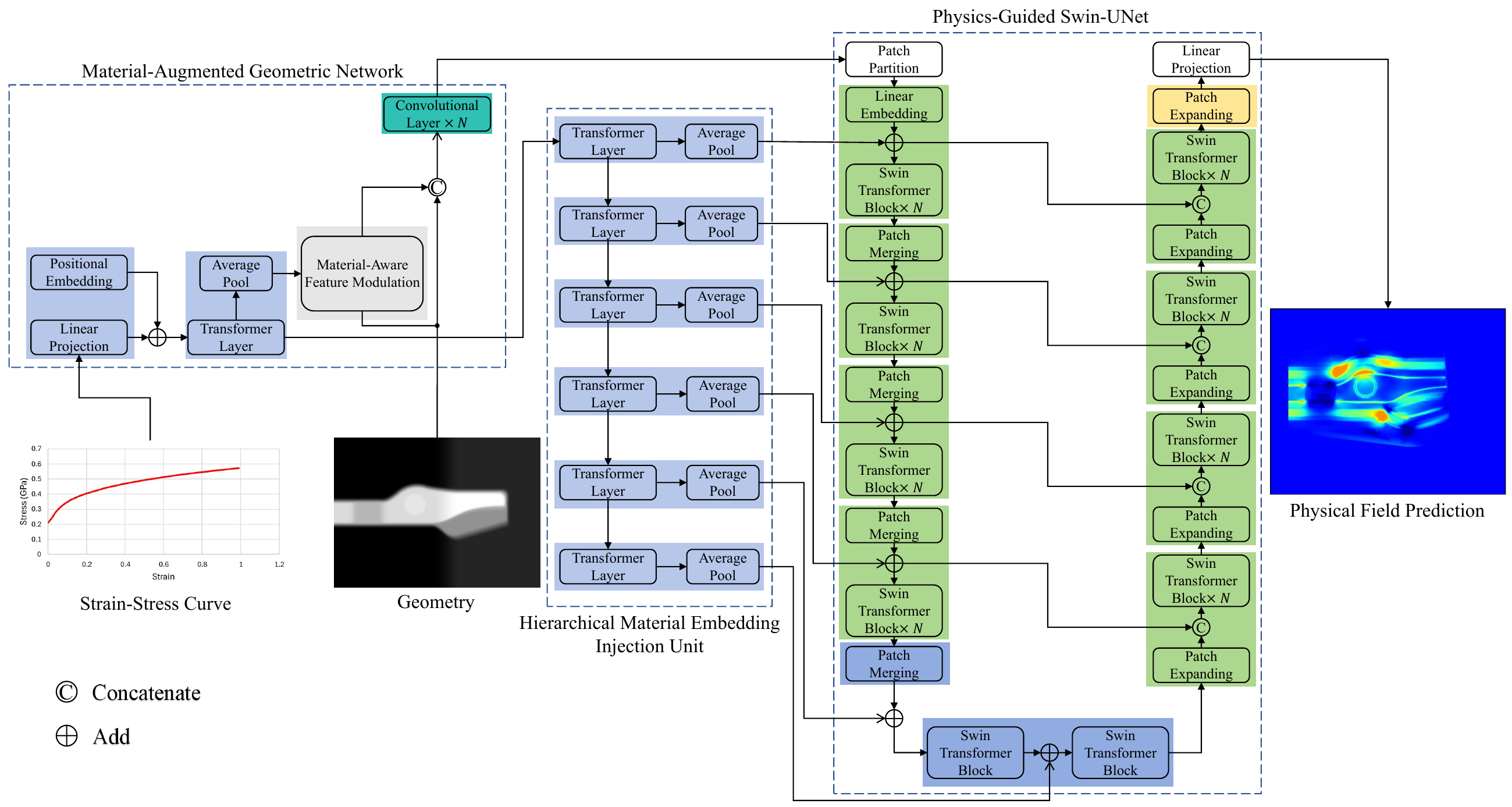}
            \caption{Overall architecture of the StampFormer model. The framework consists of three main components: the MAGN for fusing geometry and material data, the HMEIU for multi-scale feature integration, and the PGSU backbone for predicting the final physical field.}\label{fig:main_architecture}
        \end{figure}
    \end{landscape}
    \clearpage 
}

\subsection{Material-Augmented Geometric Network}
Inspired by heterogeneous data processing architectures, the MAGN in StampFormer is responsible for the initial multi-modal fusion of the 2D part geometry and the 1D stress-strain curve. This robust multi-modal integration is achieved through three key stages: material processing, material-aware feature modulation, and spatial feature integration.

For the material path, the 1D stress-strain curve $S \in \mathbb{R}^{T \times 1}$ is transformed into a dense representation. It is first linearly projected into a $d_{\rm mat}$-dimensional space using a weight matrix $W_{\rm S} \in \mathbb{R}^{1 \times d_{\rm mat}}$ and combined with a learnable positional embedding $E_{\rm pos,mat} \in \mathbb{R}^{T \times d_{\rm mat}}$. This sequence is then processed by a Transformer Encoder Layer to capture the intricate constitutive relationships within the material. Next, the output tokens from this Transformer are aggregated via average pooling (applied over the $T$ token dimension) to produce a condensed, global material embedding $E_{\rm mat} \in \mathbb{R}^{d_{\rm mat}}$:
\begin{equation}
E_{\rm mat} = \text{AvgPool}(\text{TransformerEncoder}(S W_{\rm S} + E_{\rm pos,mat})).
\end{equation}

In the Material-Aware Feature Modulation (MAFM) stage, we map the global material characteristics into the spatial geometric domain. The input geometry is represented as a single-channel image $I_{\rm geo} \in \mathbb{R}^{H \times W \times 1}$. To align the feature spaces, $E_{\rm mat}$ is passed through a multi-layer perceptron (MLP) to form a conditional material vector $\text{MLP}(E_{\rm mat}) \in \mathbb{R}^C$, while $I_{\rm geo}$ is independently processed by a point-wise convolution ($1 \times 1$ Conv) to extract its base spatial embeddings. The material vector is then spatially broadcast across the spatial dimensions and added to the geometric embeddings. This modulated feature is refined by a sequence of convolutional layers to yield an intermediate geometry-material fused feature map $F_{\rm fused} \in \mathbb{R}^{H \times W \times C}$:
\begin{equation}
F_{\rm fused} = \text{ConvLayers}(\text{Conv}_{1 \times 1}(I_{\rm geo}) + \mathcal{B}(\text{MLP}(E_{\rm mat}))),
\end{equation}
where $\mathcal{B}(\cdot)$ denotes the spatial broadcasting operation that broadcasts the vector to $\mathbb{R}^{H \times W \times C}$. This guarantees that the geometric representations are uniformly conditioned by the macroscopic material properties.

Finally, in the spatial feature integration stage, we employ a concatenation-based skip connection to preserve pristine geometric boundary information. The initial raw geometry $I_{\rm geo}$ is concatenated with the fused feature $F_{\rm fused}$ along the channel dimension:
\begin{equation}
Z_{\rm concat} = \text{Concat}(I_{\rm geo}, F_{\rm fused}).
\end{equation}
This concatenated representation seamlessly bridges the explicit physical boundaries with the deep, material-conditioned features. Subsequently, $Z_{\rm concat}$ is processed through a final integration block (denoted as $\text{ConvInt}$) consisting of consecutive $3 \times 3$ convolutional layers coupled with ReLU activations. This yields the final, high-fidelity material-augmented feature map $X^{(0)} = \text{ConvInt}(Z_{\rm concat})$, which is fed directly into the main Swin-UNet backbone.

\subsection{Physics-Guided Swin-UNet}
As the main backbone of StampFormer, the PGSU is responsible for processing the fused multi-modal feature map $X^{(0)}$ to predict the final physical field. It adapts the powerful Swin-UNet architecture, which follows a symmetric encoder-decoder structure built upon a series of Swin-Transformer blocks.

\begin{figure}[!htbp] 
\centering
\includegraphics[width=0.3\columnwidth]{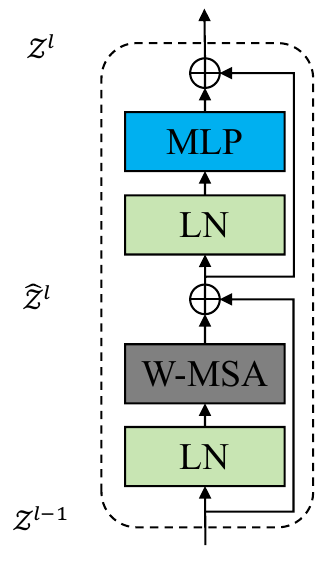}
\caption{The structure of a Swin Transformer block, which consists of Layer Normalization (LN), Windowed Multi-head Self-Attention (W-MSA), and a Multi-Layer Perceptron (MLP), with residual connections.}\label{fig:swin_block}
\end{figure}

As detailed in Figure \ref{fig:swin_block}, the core of this architecture leverages the Window-based Multi-head Self-Attention (W-MSA) mechanism. The computation for a single Swin-Transformer block operating on an input feature map $z^{l-1}$ is given by:
\begin{align}
\hat{z}^l &= \text{W-MSA}(\text{LN}(z^{l-1})) + z^{l-1}, \\
z^l &= \text{MLP}(\text{LN}(\hat{z}^l)) + \hat{z}^l,
\end{align}
where LN denotes Layer Normalization. By employing this window-based attention design, the network achieves robust feature extraction while maintaining high computational efficiency.

The encoder path takes the initial multi-modal feature map $X^{(0)}$ as its input and consists of four hierarchical stages. Each stage comprises a pair of Swin-Transformer blocks followed by a Patch Merging layer, which reduces the spatial resolution while progressively increasing the feature dimension.

The physics-guided nature of this architecture stems from its explicit modeling of the multi-scale interaction between the two core physical inputs: geometry and material properties. The physics of forming dictates that global material characteristics (a 1D concept) influence local geometric deformation (a 2D field) at all scales. Therefore, our design integrates the HMEIU (detailed in Section \ref{HMEIU}) to hierarchically inject material information at each stage of the encoder, specifically after the spatial features have been downsampled by the Patch Merging layer. This allows the model to learn how global material characteristics affect both local, fine-grained features in shallow layers, and global, abstract features in deep layers.

The decoder path is symmetric, using Patch Expanding layers to upsample the features. Crucially, at each upsampling step, the decoder path brings back high-resolution, fine-grained features from the corresponding encoder stage via skip connections. These early features are fused with the abstract, semantically rich deep features from the deeper decoder layers. This fusion ensures the final prediction is both semantically rich and spatially precise.

\begin{figure*}[!htbp]
\centering
\includegraphics[width=\textwidth]{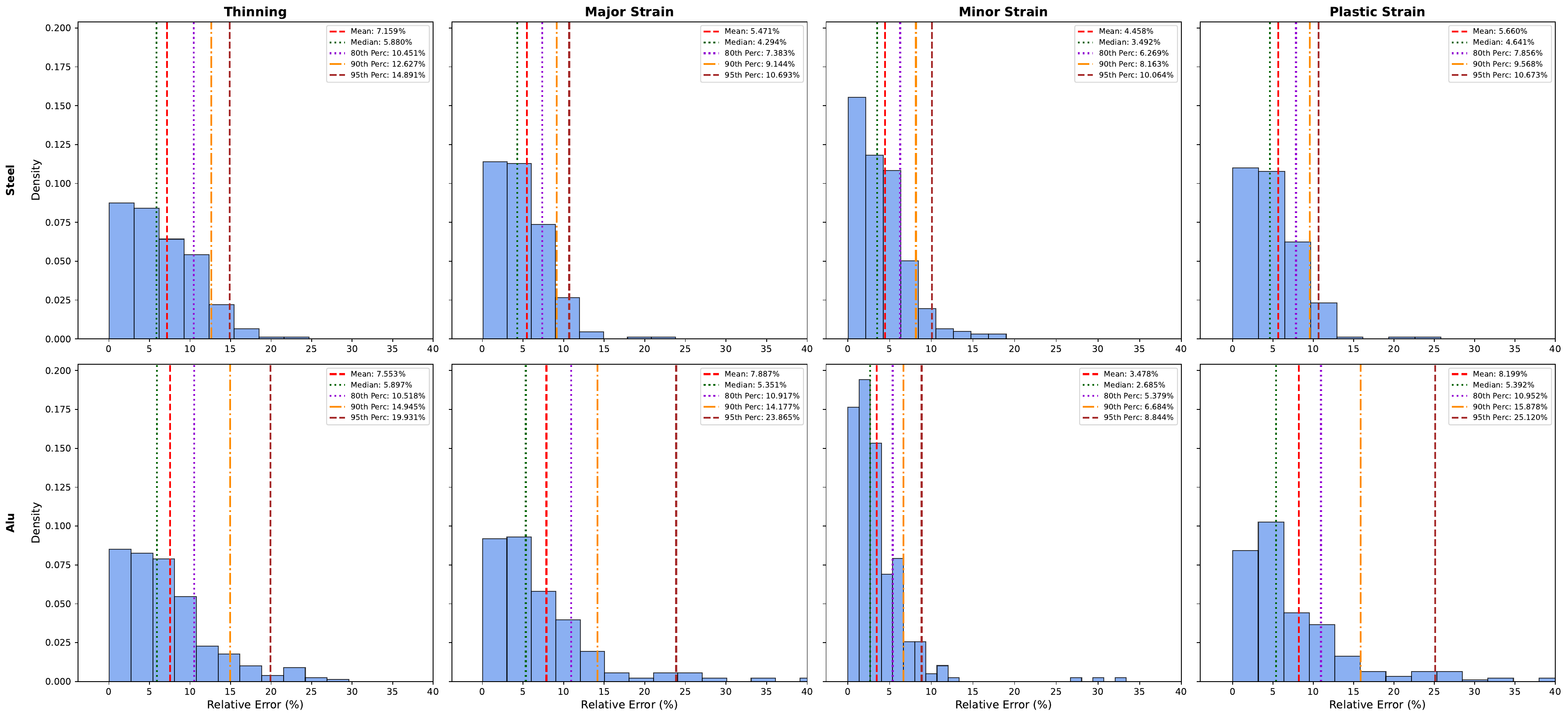}
\caption{Histograms showing the distribution of RE computed from the representative maximum, defined as the average of the top 0.1\% valid values in four 2D physical fields (i.e. thinning, major strain, minor strain and plastic strain). Additional lines are introduced to mark the mean, median, and various percentiles.}\label{fig:relative_error_dist}
\end{figure*}
\begin{figure}[!htbp]
\centering
\includegraphics[width=\linewidth]{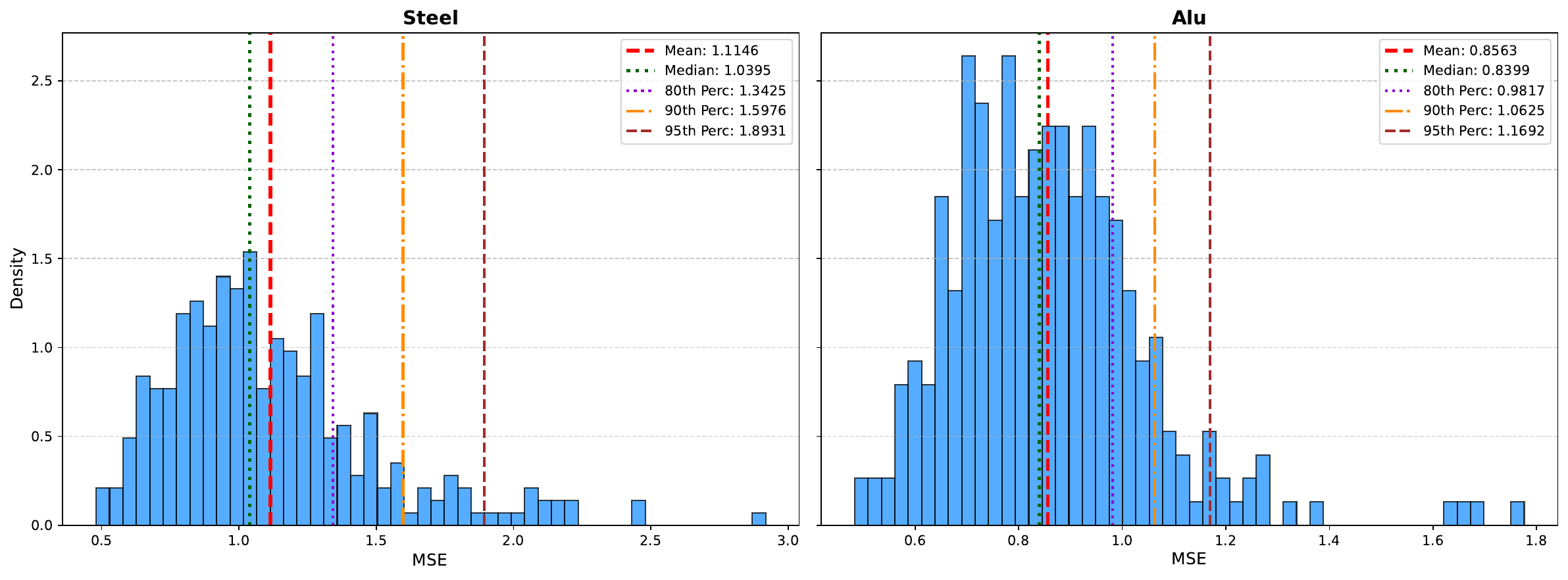}
\caption{Histograms showing the distribution of MSE (mm$^2$) for the 3D displacement field. Additional lines are introduced to mark the mean, median, and various percentiles.}\label{fig:mse_dist}
\end{figure}

\subsection{Hierarchical Material Embedding Injection Unit} \label{HMEIU}
Instead of a single fusion at the MAGN, we introduce the HMEIU to ensure a deep, multi-scale integration of material embeddings throughout the network's main backbone, PGSU. It provides material embeddings of appropriate size at each stage of the feature extraction hierarchy, allowing the model to leverage material properties in conjunction with geometric features at various scales.

As shown in Figure \ref{fig:main_architecture}, the HMEIU operates as a parallel tower to the PGSU's down-sampling encoder. It takes the sequence of token embeddings from the initial stress-strain curves Transformer in the MAGN as its input, and is then progressively processed through a stack of dedicated Transformer Layers.

At each hierarchical level $l$ of the PGSU encoder, the corresponding Transformer Layer within the HMEIU refines the stress-strain curves. Crucially, each of these dedicated Transformer layers is designed to output a feature dimension that perfectly aligns with the channel dimension of the corresponding PGSU encoder stage. The output tokens from this layer are then aggregated via average pooling to produce the level-specific material embedding, $E_{\rm mat}^{(l)}$. As shown in Figure \ref{fig:main_architecture}, this embedding is injected into the main data stream of the PGSU after the Swin-Blocks and Patch Merging layer at each encoder stage. Let $X^{(l-1)}$ be the input to the $l$-th stage. The geometric features are first processed by the two consecutive Swin Transformer blocks and then downsampled by the Patch Merging layer to obtain the intermediate spatial feature $F^{(l)} \in \mathbb{R}^{N_l \times C_l}$:
\begin{equation}
F^{(l)} = \text{PatchMerging}(\text{SwinBlock}(\text{SwinBlock}(X^{(l-1)}))).
\end{equation}
$E_{\rm mat}^{(l)}$ is then broadcast and added to geometric features using an outer-product formulation to match dimensions:
\begin{equation}
X^{(l)} = F^{(l)} + \mathbf{1}_{N_l} (E_{\rm mat}^{(l)})^T,
\end{equation}
where $\mathbf{1}_{N_l} (E_{\rm mat}^{(l)})^T \in \mathbb{R}^{N_l \times C_l}$ and $\mathbf{1}_{N_l}$ represents an $N_l$-dimensional column vector of ones, explicitly denoting the spatial broadcasting of the material embedding across all geometric patches. This hierarchical injection mechanism ensures that as the geometric features become more abstract and complex during the downsampling process, the material representation is co-adapted and fused at each corresponding scale.

\section{Experiments}
\subsection{Training Setup}
In this study, a total of ten separate models were trained: one dedicated StampFormer model for each of the five target physical fields (thinning, major strain, minor strain, plastic strain, and displacement), for each of the two material datasets (steel and aluminium). All ten models were trained using a consistent experimental setup on their respective data, as described in Section \ref{dataprep}. We employed the Adam optimizer for model training with standard parameters: beta parameters of $\beta_1 = 0.9$ and $\beta_2 = 0.999$ and an epsilon of $\epsilon = 10^{-8}$. The learning rate was managed by a step learning-rate scheduler, with an initial rate of $10^{-4}$. The scheduler was configured to decrease the learning rate by a factor of $\gamma = 0.4$ every $100$ epochs. The optimization goal for each model was to minimize the Mean Squared Error (MSE) between its predicted physical field and the ground truth.

Considering the predicted field $\hat{Y} \in \mathbb{R}^{H \times W \times C}$ and the ground truth field $Y \in \mathbb{R}^{H \times W \times C}$ from the FEA simulation, the MSE loss is computed on the physical units and defined as:
\begin{equation}
\mathcal{L}_{\rm MSE} = \frac{1}{H \cdot W \cdot C} \sum_{i=1}^{H} \sum_{j=1}^{W} \sum_{k=1}^{C} (Y_{ijk} - \hat{Y}_{ijk})^2.
\end{equation}
Here, $H$ and $W$ are the height and width of the field, and $C$ is the number of channels. Note that $C=1$ for the four 2D fields, and $C=3$ for the 3D displacement field.

\subsection{Metrics}
\label{exp.metrics}

In addition to MSE, we also employ additional metrics that are more meaningful from an engineering perspective to evaluate the trained models on the four 2D physical fields: thinning, major strain, minor strain, and plastic strain. For these fields, we focused on the accuracy of predicting the maximum value, as this is often of critical importance in forming analysis. To ensure stability and mitigate the effect of potential outliers, we define the representative maximum value as the average of the top $0.1\%$ of values in a given field. Let $m_{0.1}(Y)$ and $m_{0.1}(\hat{Y})$ be the representative maximum values for the ground truth and predicted fields, respectively. We then calculate the Relative Error (RE) as:
\begin{equation}
\label{rele}
\text{Relative Error (\%)} = 100 \times \frac{|m_{0.1}(Y) - m_{0.1}(\hat{Y})|}{m_{0.1}(Y)}.
\end{equation}

\begin{figure*}[!htbp]
\centering
\includegraphics[width=\textwidth]{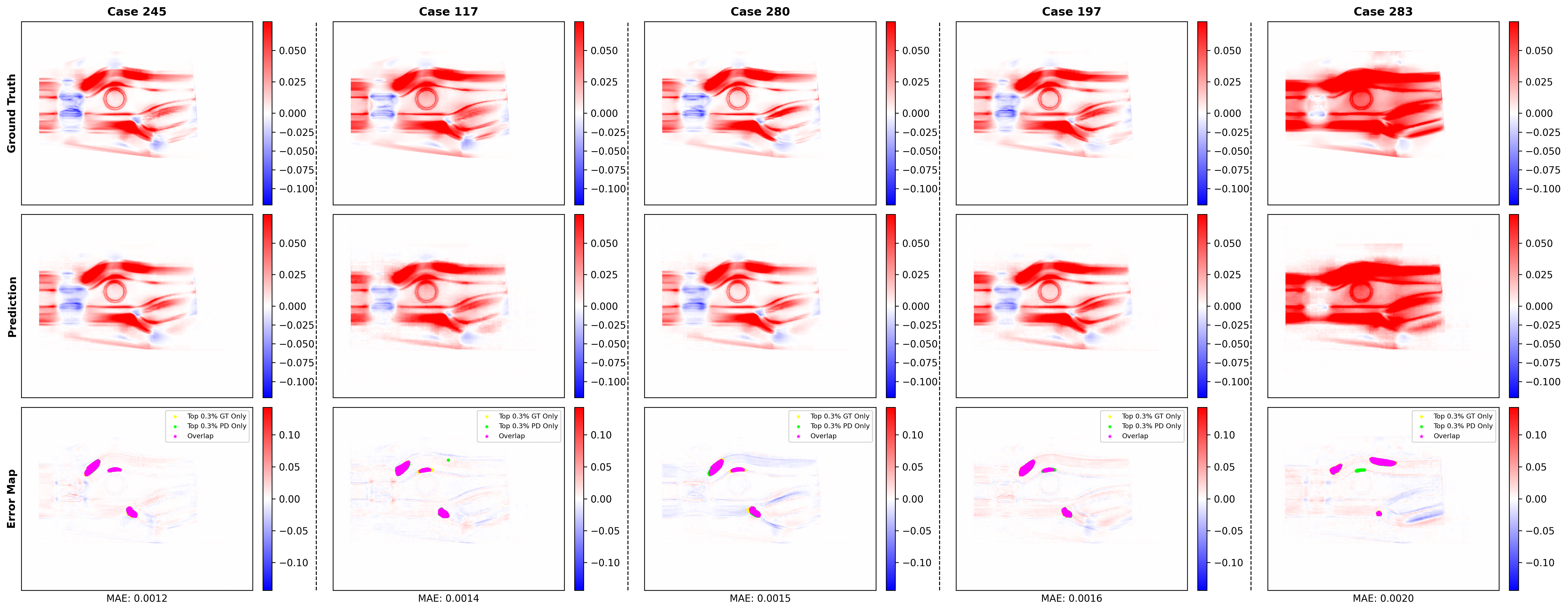}
\caption{Visualization for thinning field prediction on five representative cases from the steel dataset. The top-value overlap map specifically visualizes the top 0.3\% of values to compare the prediction of maximum thinning, highlighting regions of overlap, GT Only, and PD Only (note that 0.3\% is used for visualization, whereas 0.1\% is used for the quantitative metric).}\label{fig:steel_thinning}
\end{figure*}
\begin{figure*}[!htbp]
\centering
\includegraphics[width=\textwidth]{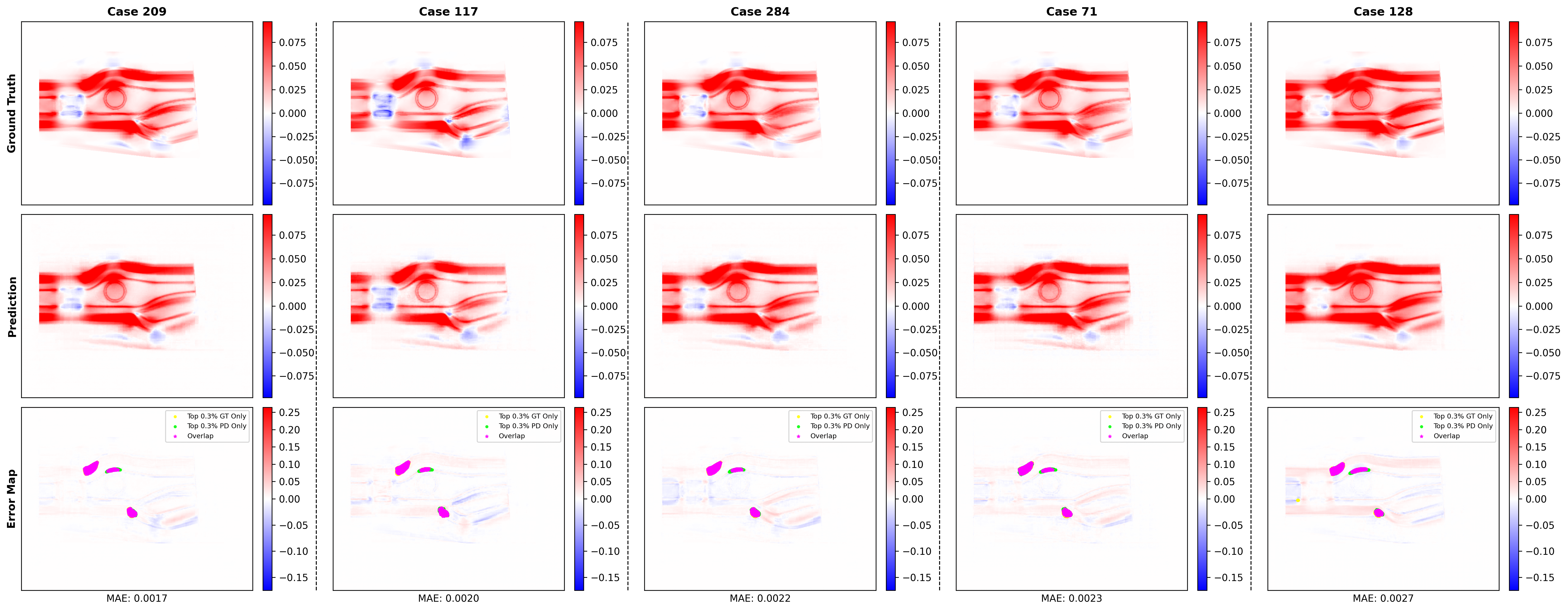}
\caption{Visualization for thinning field prediction on five representative cases from the aluminium dataset. The top-value overlap map specifically visualizes the top 0.3\% of values to compare the prediction of maximum thinning, highlighting regions of overlap, GT Only, and PD Only (note that 0.3\% is used for visualization, whereas 0.1\% is used for the quantitative metric).}\label{fig:alu_thinning}
\end{figure*}
\begin{figure*}[!htbp]
\centering
\includegraphics[width=\textwidth]{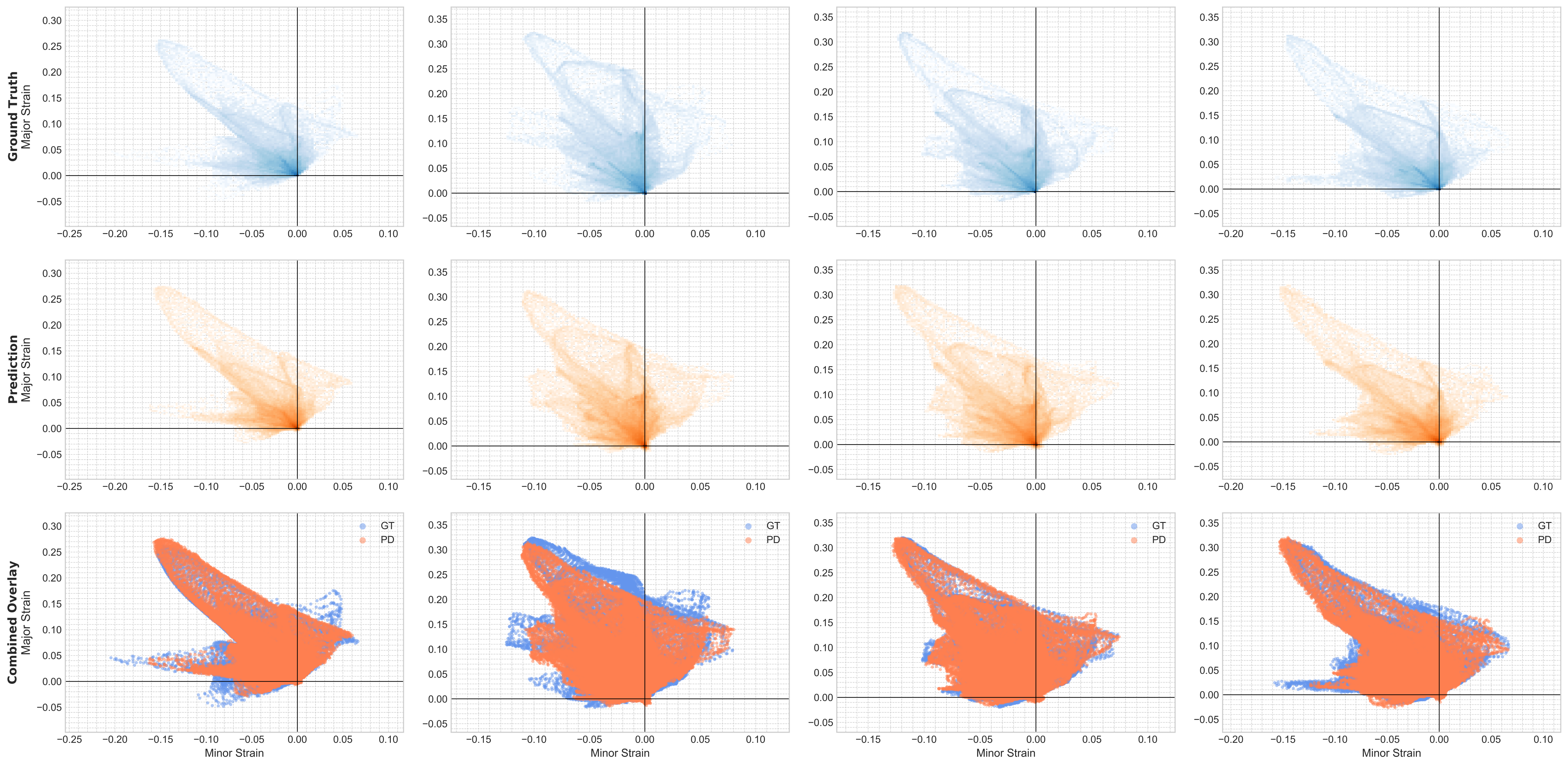}
\caption{Comparison of FLDs from the steel dataset. The top two rows display the ground truth and predicted FLDs, respectively. For direct comparison, the bottom row provides an overlay of both.}\label{fig:steel_fld}
\end{figure*}
\begin{figure*}[!htbp]
\centering
\includegraphics[width=\textwidth]{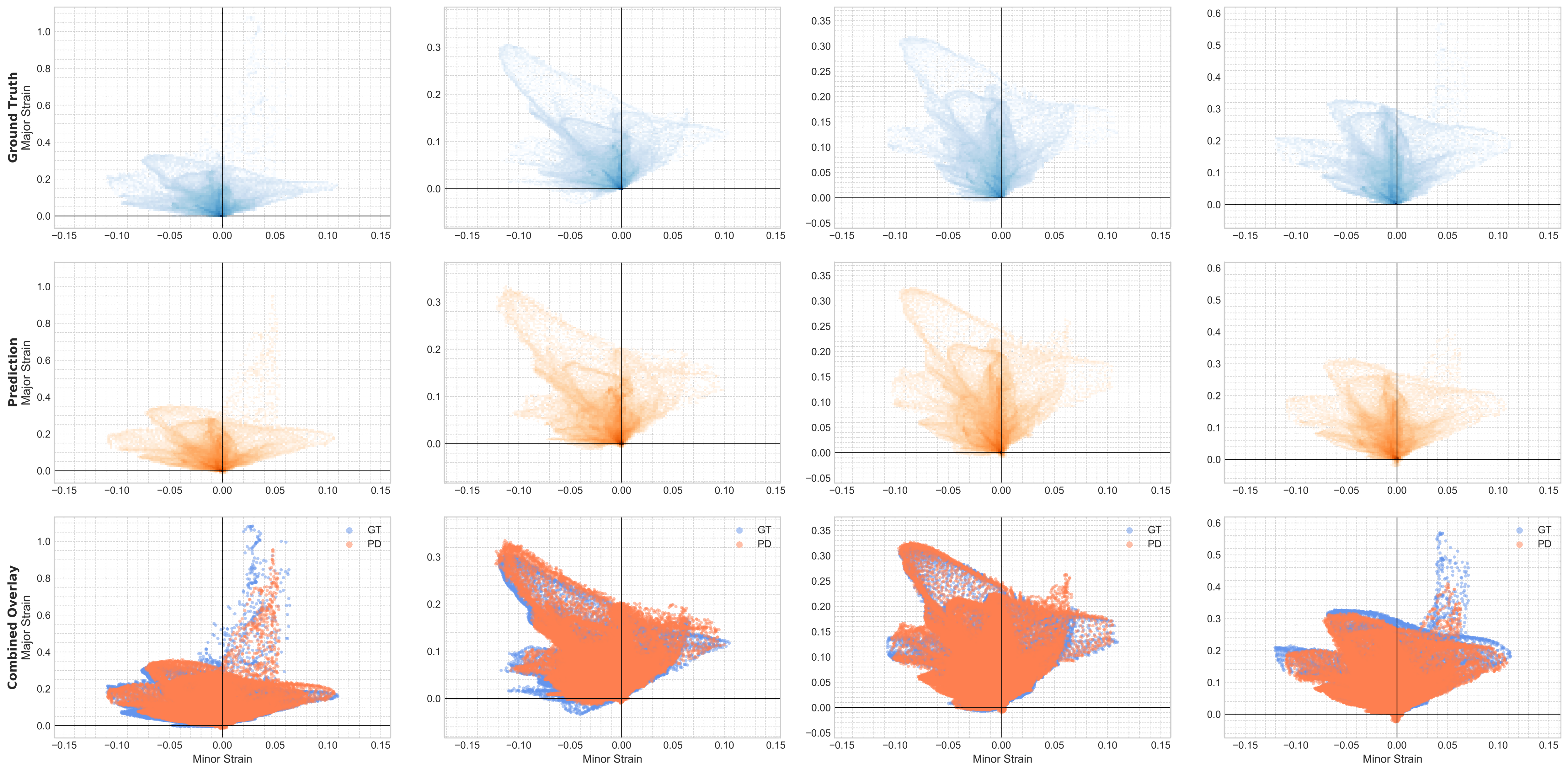}
\caption{Comparison of FLDs from the aluminium dataset. The top two rows display the ground truth and predicted FLDs, respectively. For direct comparison, the bottom row provides an overlay of both.}\label{fig:alu_fld}
\end{figure*}
\begin{figure*}[!htbp]
\centering
\includegraphics[width=\textwidth]{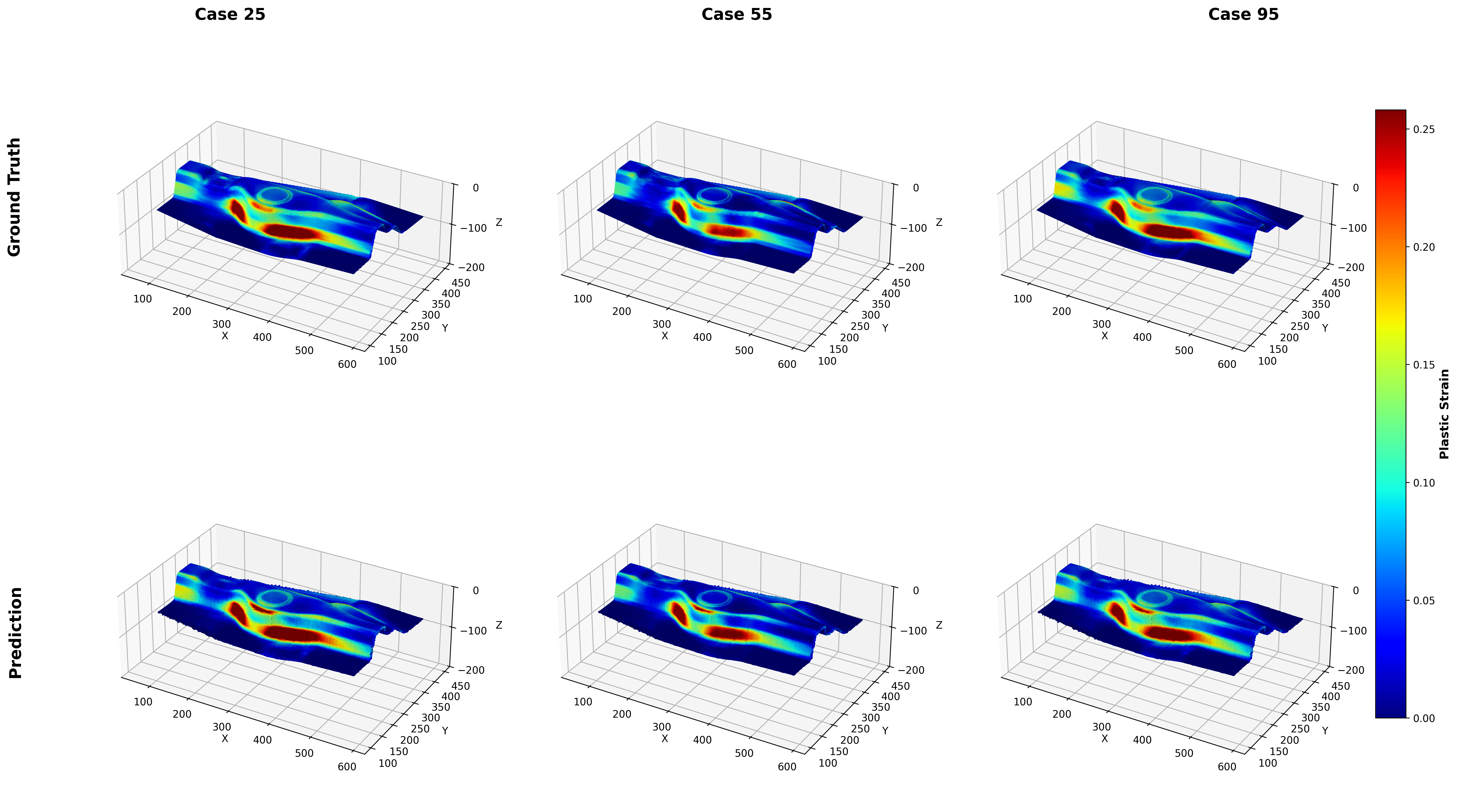}
\caption{Visualization of displacement-reconstructed surfaces colored by plastic strain from the steel dataset. The comparison between GT and PD highlights the model's accuracy in predicting the shape and plastic strain distribution of the final stamping part.}\label{fig:steel_displastic}
\end{figure*}
\begin{figure*}[!htbp]
\centering
\includegraphics[width=\textwidth]{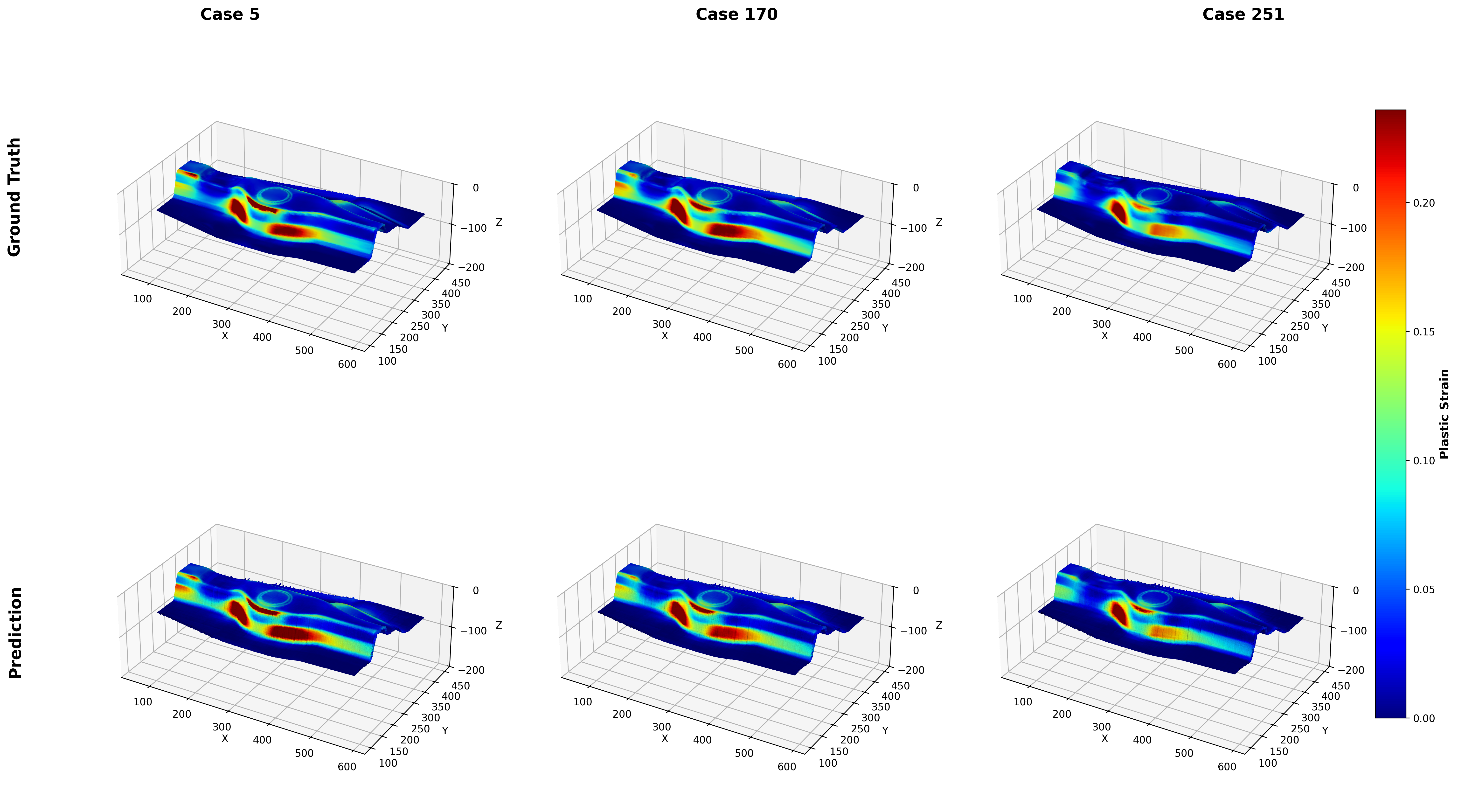}
\caption{Visualization of displacement-reconstructed surfaces colored by plastic strain from the aluminium dataset. The comparison between GT and PD highlights the model's accuracy in predicting the shape and plastic strain distribution of the final stamping part.}\label{fig:alu_displastic}
\end{figure*}

\section{Results and Discussion}
In this section, we present a comprehensive evaluation of the StampFormer model. We begin with a detailed quantitative analysis of the prediction accuracy using numerical metrics. The following sections offer a visual analysis of the model's performance by presenting the predicted physical fields, demonstrating the qualitative agreement with the FEA ground truth.

\subsection{Quantitative Analysis}
In this section, we perform a detailed quantitative analysis to evaluate the performance of StampFormer. To begin with, the MSE on the test set for both the steel and aluminium datasets across five different physical fields, i.e. thinning, major strain, minor strain, plastic strain, and displacement, was calculated. The results are summarized in Table \ref{tab:mse_results}, and Table \ref{tab:mse_results} shows that the model achieves low MSE values for the 2D physical fields on both datasets, indicating a high degree of prediction accuracy. It is worth noting that, as shown in Figure \ref{fig:steel_displastic} and Figure \ref{fig:alu_displastic}, the displacement field is a three-component field with a larger numerical range, which makes a higher MSE reasonable. In general, the model demonstrates strong performance based on the MSE on the test set.

\begin{table}[!htbp] 
\centering
\caption{Test Set MSE for Different Physical Fields.}
\label{tab:mse_results}
\begin{tabularx}{\columnwidth}{@{}l *{5}{>{\centering\arraybackslash}X} @{}}
\toprule
\textbf{Dataset} & \textbf{Thinning} & \textbf{Major Strain} & \textbf{Minor Strain} & \textbf{Plastic Strain} & \textbf{Displacements (mm$^2$)} \\
\midrule
Steel & 2.64E-05 & 3.87E-05 & 7.78E-05 & 7.20E-05 & 0.605 \\
Aluminium & 4.37E-05 & 1.48E-04 & 5.17E-05 & 9.62E-05 & 0.342 \\
\bottomrule
\end{tabularx}
\end{table}

\begin{figure}[!htbp]
\centering
\includegraphics[width=0.6\linewidth]{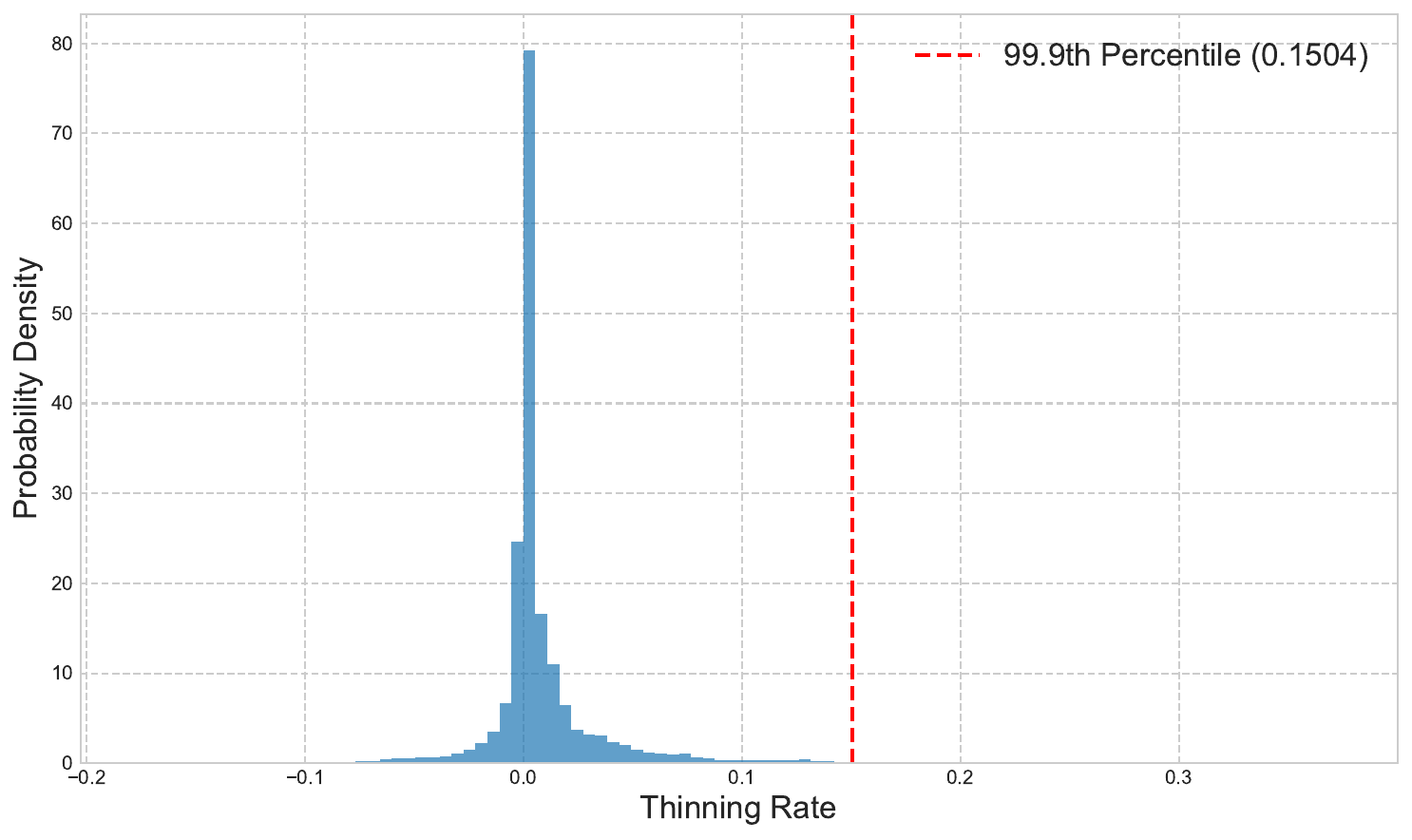}
\caption{The thinning field distribution of a random sample (where negative values indicate thickening), with a vertical line indicating the 99.9th percentile (top 0.1\%) of values.}
\label{fig:thinning_distribution_percentile}
\end{figure}

\begin{figure*}[!htbp]
    \centering
    \begin{subfigure}[b]{0.32\linewidth} 
        \centering
        \includegraphics[width=\textwidth]{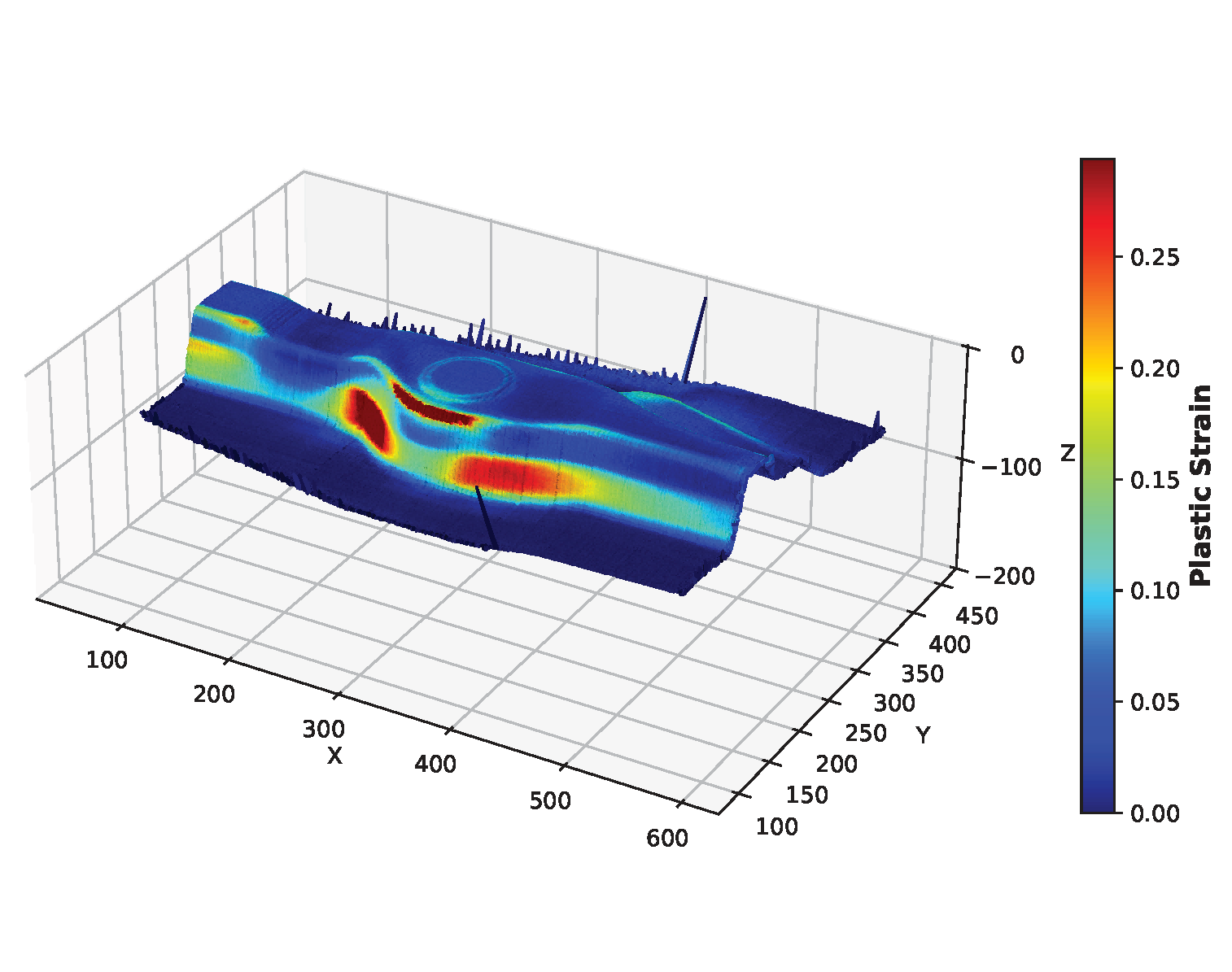}
        \caption{Prediction before de-noising}
        \label{fig:dis_before}
    \end{subfigure}
    \hfill 
    \begin{subfigure}[b]{0.32\linewidth}
        \centering
        \includegraphics[width=\textwidth]{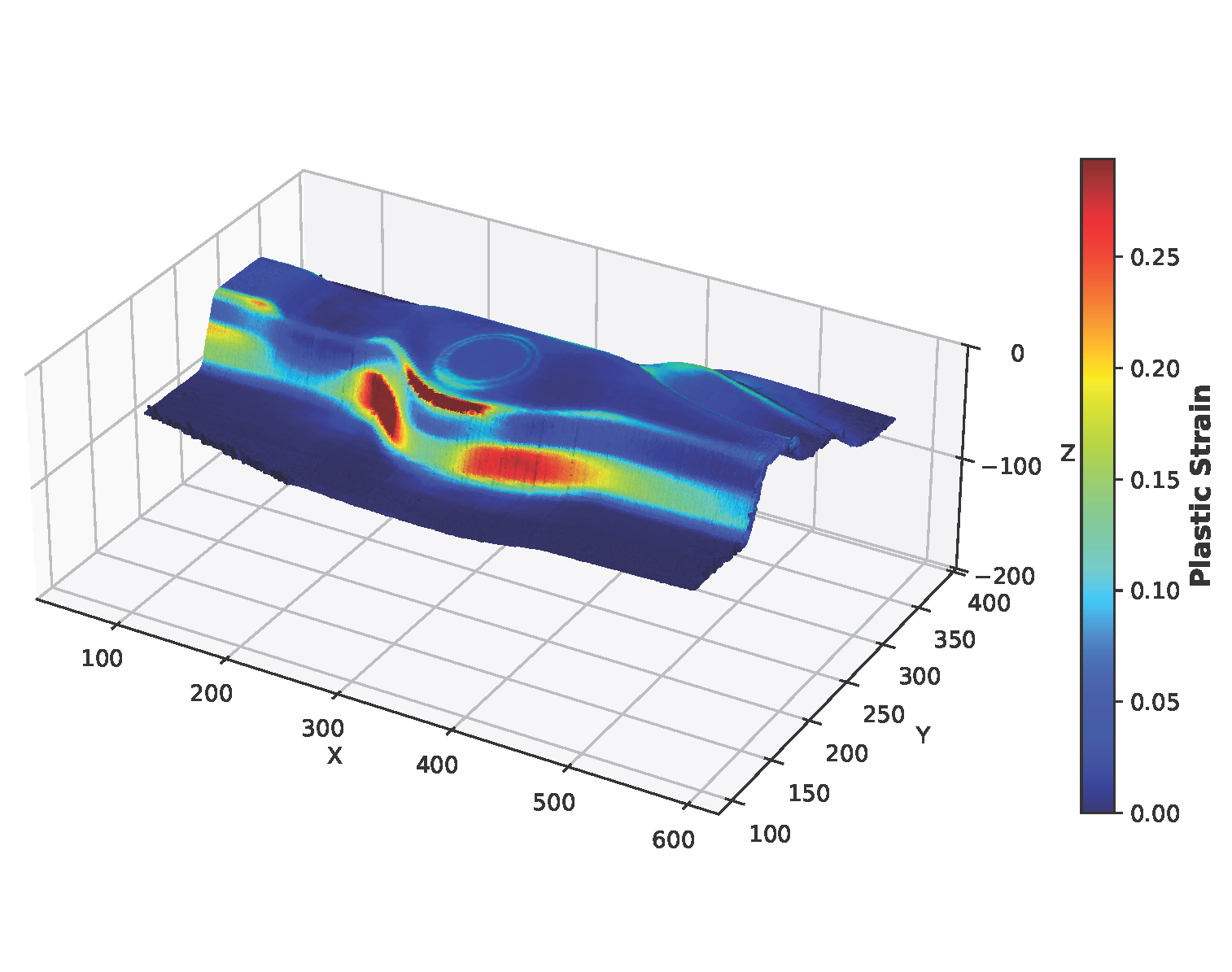}
        \caption{Prediction after de-noising}
        \label{fig:dis_after}
    \end{subfigure}
    \hfill
    \begin{subfigure}[b]{0.32\linewidth}
        \centering
        \includegraphics[width=\textwidth]{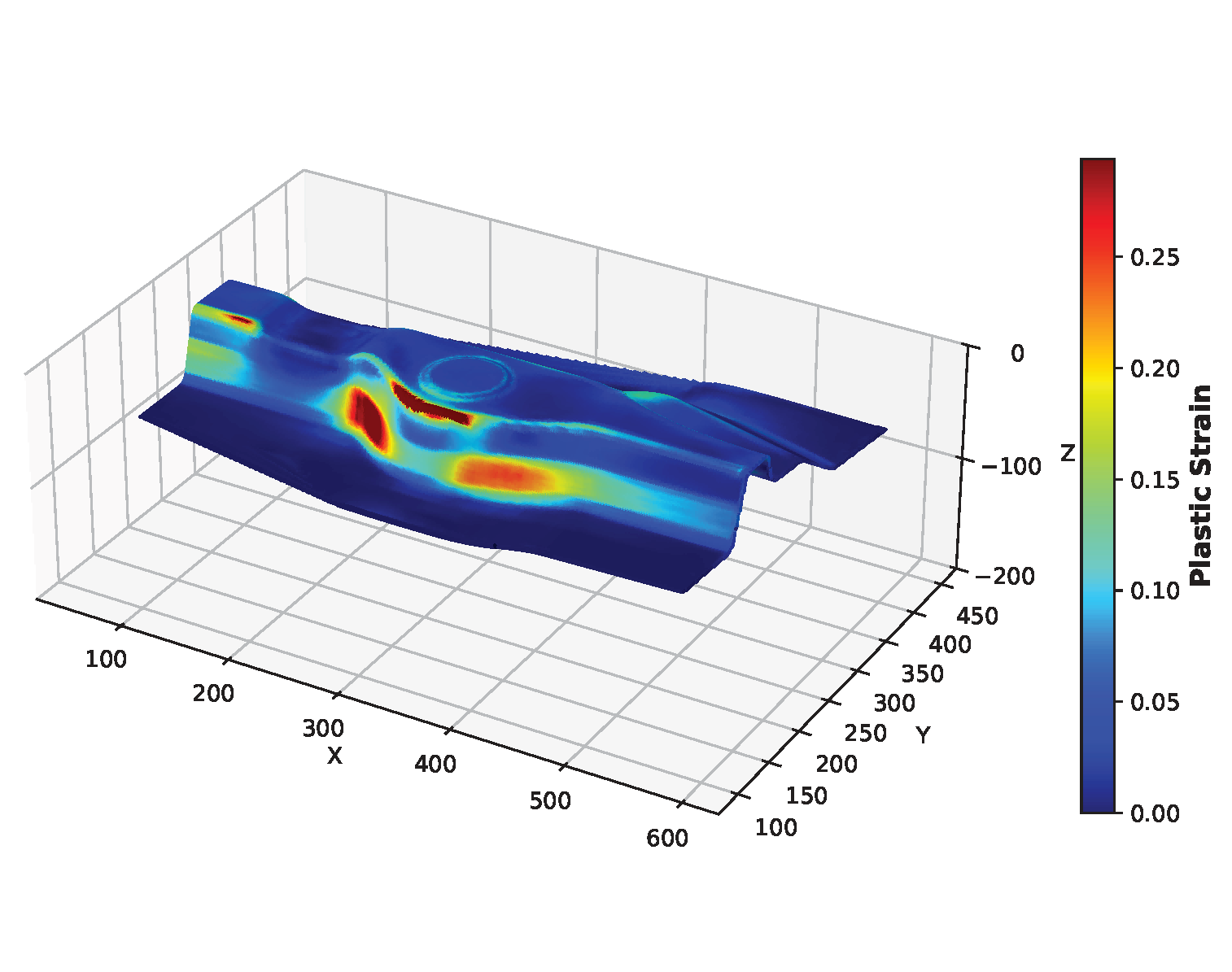}
        \caption{Ground Truth}
        \label{fig:dis_true}
    \end{subfigure}
    \caption{Comparison of surfaces reconstructed from displacement and colored by plastic strain; denoising is used only for visualization of boundary artifacts caused by zero-padding in the training data.}
    \label{fig:denoising_comparison}
\end{figure*}

We recognize the limitations of using only MSE to evaluate model performance. Therefore, we also assess the model with the descriptive metrics from Section \ref{exp.metrics}. For four 2D physical fields, performance is evaluated using the RE as defined in Equation \eqref{rele}. We illustrate the distribution of RE values for four 2D physical fields in Figure \ref{fig:relative_error_dist}. For the steel dataset, the mean RE is consistently low across all four fields, ranging from 4.5\% to 7.2\%, with the 95th percentile below 15\% in most cases. This indicates that the model's predictions for the most critical regions, i.e. areas of maximum strain/thinning, are highly accurate. 

The aluminium data showed slightly higher errors, with means ranging from 3.5\% to 8.2\%. For both materials, the error distributions were right-skewed, confirming that significant errors were uncommon. However, the aluminium dataset exhibits heavier tails, particularly for major and plastic strain, where the 95th percentile errors are substantially larger. This is likely due to the inherently lower ductility of aluminium compared to steel, which causes sharper, more localized strain gradients that are more challenging for the surrogate model to capture consistently. 

In addition, we present the distribution of MSE for the 3D displacement field in Figure \ref{fig:mse_dist}. For steel, the MSE distribution was slightly wider with a mean of 1.115 mm$^2$, while the aluminium predictions were more tightly clustered around a mean of 0.856 mm$^2$. Despite this difference, both results demonstrate a high level of accuracy, confirming the model's ability to precisely predict the final 3D part shape.


\subsection{Thinning Field Prediction}
In this section, we illustrate the results for the thinning field in Figures \ref{fig:steel_thinning} and \ref{fig:alu_thinning}. Each case is presented in three rows: the ground truth field from FEA, the corresponding prediction from our model, and a detailed top-value overlap map along with the prediction error. In industrial applications, the prediction of maximum thinning is of great value. To better evaluate the prediction of maximum thinning, we visualize the regions with the highest values on the top-value overlap map. Specifically, we plot the top 0.3\% of thinning values from both the ground truth (GT) and the prediction (PD). The top-value overlap map uses a color-coded system to distinguish these regions: yellow highlights areas where high values are found only in the ground truth, green marks areas exclusive to the prediction, and magenta indicates where they overlap. Building on our earlier quantitative analysis of the maximum thinning, the significant overlap observed in the top-value overlap maps for both datasets demonstrates the model's high accuracy in identifying both the location and magnitude of critical thinning regions.

\subsection{Forming Limit Diagram Prediction}
The Forming Limit Diagram (FLD) is a fundamental tool for evaluating sheet metal formability by representing the critical combinations of major and minor strains at the onset of localized necking or fracture during sheet metal forming. In an FLD, the horizontal axis represents the minor strain, while the vertical axis corresponds to the major strain. In this section, we compare the predicted FLDs with the ground truth for both datasets in Figures \ref{fig:steel_fld} and \ref{fig:alu_fld}. In these figures, the top two rows display the ground truth and predicted FLDs, respectively, where the density of the data points is clearly visible. For a direct comparison, the last row overlays the predicted points (orange) onto the ground truth points (blue). The close correspondence in the shape, density, and spread of the point clouds in the overlay plots indicates that our model accurately captures the complex formability characteristics under different conditions.

\subsection{Displacement and Plastic Strain Prediction}
\label{displacementvis}
As shown in Figures \ref{fig:steel_displastic} and \ref{fig:alu_displastic}, we compare the predicted and ground-truth as-formed components, which are reconstructed from the displacement fields. On these surfaces, we superimpose the corresponding plastic strain fields. The top row in each figure shows the ground truth from the FEA simulation, while the bottom row shows our model's prediction. Because the model's predictions can be affected by the zero-padding values at the edges, some noise may appear. To enhance the clarity of the 3D renderings of the displacement fields, we applied a localized de-noising process as a visual post-processing step. The visual comparison reveals a strong agreement in both the final shape of the stamping parts and the plastic strain field for both steel and aluminium datasets. This confirms the model's capability to predict displacement and plastic strain with high fidelity.

\section{Discussion}

\subsection{Engineering Significance of the Proposed Method}
In the traditional design cycle, a primary bottleneck is that time-consuming and computationally expensive FEA hinders rapid design iteration and optimization. In this study, we introduced StampFormer, a surrogate model designed to predict complex physical fields in the sheet metal forming process with high accuracy and high speed. To quantify this, each FEA simulation in our dataset required several hours of computation to complete. In stark contrast, StampFormer provides predictions for physical fields in under a second on a single GPU. This acceleration, representing a speed-up of more than three orders of magnitude (from hours to sub-second), directly enables the real-time feedback that was previously impossible.

Beyond this speed-up, to the best of the authors’ knowledge, StampFormer is the first surrogate model that explicitly incorporates both geometry and material properties. It shows a level of generalization that is essential for practical industrial use, where material selection is a key design variable. Furthermore, our model goes beyond previous approaches by predicting a wider range of physical fields, providing a more complete picture of the forming process. We demonstrate its capability by accurately predicting critical thinning regions, full FLD, plastic strain distributions, and the final 3D shape of the part. This offers design engineers a much richer and more informative assessment.

\subsection{Discussion on the Relative Error Metric}
In Section \ref{exp.metrics}, we defined a field-specific relative-error metric, which uses a representative maximum value, calculated from the average of the top 0.1\% of values in a given field, rather than the absolute maximum value. This decision was made to enhance the numerical stability and robustness of our evaluation metric. As illustrated in Figure \ref{fig:thinning_distribution_percentile}, the distribution of values in a physical field like thinning can have a long tail, where a small number of nodes may exhibit extreme values due to numerical instability. Relying on a single maximum value can make the metric highly sensitive to these outliers. As a result, we average the top 0.1\% of values to obtain a more stable and representative measure of the model's performance in the most critical regions, which is of greater practical importance in an industrial context.

\subsection{Rationale for De-noising Displacement Field Predictions}
As mentioned in Section \ref{displacementvis}, we applied a de-noising step to the predicted displacement fields. The valid domain of the FEA output is non-rectangular. To create a uniform input for the neural network, we pad the non-rectangular simulation results with zero values to fit them into a standard rectangular image format. While this is a common practice, the sharp discontinuity between the actual data and the padded background regions can introduce instability at the edges of the model's predictions, resulting in noisy artifacts, as shown in Figure \ref{fig:denoising_comparison}. After applying a de-noising process that involves removing irrelevant fringe artifacts and using Gaussian de-noising, we effectively eliminate these artifacts. Because the effect of this filtering is strictly confined to removing numerical noise at the artificial boundaries, it serves as a highly effective post-processing step for visualization. Consequently, it cleanly recovers the true contours of the part, greatly enhancing the visual quality and readability of the final 3D shape.

\section{Conclusion}
This paper presents StampFormer, a physics-guided, material–geometry-coupled multimodal framework for rapid prediction of forming-response fields in sheet metal stamping. By integrating 2D component geometry with 1D material-response information, the model captures the influence of material behaviour on spatial deformation, thinning and strain evolution. The proposed architecture combines the Material-Augmented Geometric Network, the Hierarchical Material-Embedded Injection Unit and a Physics-Guided Swin-UNet backbone to embed material information across multiple feature scales while preserving field-level spatial detail.

The results demonstrate that StampFormer predicts key forming outputs with high accuracy, including thinning, plastic strain, major-minor strain distributions and 3D displacement fields. The predicted fields show strong agreement with finite-element simulation results in both quantitative metrics and visual comparisons. The model achieves an average relative error below 8.5\% for the 2D forming fields and a displacement prediction MSE error of 1.2 mm$^2$, indicating its potential as an efficient surrogate for high-cost finite-element simulations.

The investigated steel and aluminium property ranges span an industrially relevant cold-forming material design envelope, covering the principal response characteristics of commonly used forming alloys. Together with data augmentation, this enables StampFormer to learn a broad range of material-dependent forming responses within the defined design domain. Although transferability to entirely new part families remains an important direction for future work, the current framework establishes a scalable basis for material–geometry–process screening. Overall, StampFormer offers a promising route towards faster forming assessment, reduced dependence on repeated simulations and more efficient early-stage design of stamped components.

\section*{Acknowledgements}
The authors would like to thank ESI Group for technical support with the PAM-STAMP software. This work was supported by the Innovate UK Smart Grant (Project Reference: 10083425) for Imperial College London and Multi-X Solution Limited, an Innovate UK KTP fund (10159356), and a start-up fund (OSR/0550/SASC/S022) for Dr Jichun Li from Newcastle University. For the purpose of open access, the authors have applied a Creative Commons Attribution (CC BY) licence to any Author Accepted Manuscript version arising from this work.

%
%

\bibliographystyle{elsarticle-num}
\bibliography{ref}

\end{document}